\documentclass{article}

 \usepackage[preprint]{neurips_2026}

\usepackage[utf8]{inputenc} 
\usepackage[T1]{fontenc}    
\usepackage{hyperref}       
\usepackage{url}            
\usepackage{booktabs}       
\usepackage{amsfonts}       
\usepackage{nicefrac}       
\usepackage{microtype}      
\usepackage{xcolor}         

\setcitestyle{round}
\usepackage{amsmath}
\usepackage{amssymb}
\usepackage{mathtools}
\usepackage{amsthm}

\usepackage{algorithm}
\usepackage{algpseudocode}

\usepackage{graphicx, multirow, makecell, colortbl}

\theoremstyle{plain}
\newtheorem{theorem}{Theorem}[section]

\newtheorem{lemma}[theorem]{Lemma}
\newtheorem{corollary}[theorem]{Corollary}
\theoremstyle{definition}

\newtheorem{assumption}[theorem]{Assumption}
\theoremstyle{remark}
\newtheorem{remark}[theorem]{Remark}

\title{\textsc{FedOAED}: Federated On-Device Autoencoder Denoiser for Heterogeneous Data under Limited Client Availability}

\author{%
  S M Ruhul Kabir Howlader \\
  School of Computing and Mathematical Sciences\\ University of Leicester \\
  Leicester LE1 7RH, United Kingdom \\
  \texttt{smrkh1@leicester.ac.uk} \\
  \And
  Xiao Chen\\  
  School of Artificial Intelligence\\ Nanchang University \\
  Nanchang 330031, China \\
  \texttt{xiao.chen@ncu.edu.cn} \\
  \And
  Yifei Xie \\
  School of Management\\ Zhejiang University \\
  Hangzhou 310027, China \\
  \texttt{xieyfzju@outlook.com} \\
  \And
  Lu Liu \\
  Department of Computer Science\\ University of Exeter \\
  Exeter EX4 4QF, United Kingdom \\
  \texttt{l.liu3@exeter.ac.uk} \\
}

\begin{document}

\maketitle

\begin{abstract}
  Federated Learning (FL) struggles with data heterogeneity and \emph{client drift}, which are severely amplified by partial-participation variance in cross-device settings. Current methods addressing this via server-side caching require $\mathcal{O}(Nd)$ memory (client population $N$ and model size $d$), which is prohibitive at scale, while stateful client-side corrections go stale due to infrequent device sampling. To address this, we propose \textsc{FedOAED} (Federated On-device Autoencoder Denoiser), a stateless, entirely client-side algorithm robust to extreme low-participation regimes. \textsc{FedOAED} trains an on-device autoencoder on a loss-weighted sliding window of the client's parameter trajectory. By mixing this reconstruction back into the local parameters, it actively filters optimisation noise and reduces update dispersion before server transmission, imposing \emph{zero} additional server memory overhead. Empirical evaluations on standard FL benchmarks under severe non-IID conditions demonstrate that \textsc{FedOAED} outperforms state-of-the-art $\mathcal{O}(Nd)$-memory baselines in most cases, achieving its largest gains under extreme heterogeneity and 1-2\% client participation. Theoretically, we prove that \textsc{FedOAED}'s gated refinement acts as a norm-bounded perturbation, preserving FedAvg's $\mathcal{O}\bigl(1/\sqrt{M\tilde{K}T}\bigr)$ non-convex stationarity rate up to an explicitly controllable neighbourhood that vanishes under an annealed mixing schedule.
\end{abstract}

\section{Introduction}\label{sec:intro}

Machine learning (ML) models heavily rely on large-scale, high-quality datasets; however, acquiring such datasets is often difficult due to privacy concerns, which leads to isolated data silos. For example, in healthcare, regulations such as the Health Insurance Portability and Accountability Act (HIPAA) and the General Data Protection Regulation (GDPR) restrict sharing of sensitive data. Federated Learning (FL) has therefore emerged as an effective approach that enables collaborative training of a global model without requiring access to participants' raw data.

The first federated learning algorithm, FedAvg, was introduced by \cite{McMahan2016}. Practical deployments are commonly divided into cross-silo and cross-device settings \cite{Kairouz2019}. In the cross-silo setting only a limited number of reliable clients (e.g., hospitals) participate in training, whereas the cross-device setting involves a very large population of often intermittently available devices (e.g., mobile phones). In cross-device FL a small, randomly sampled fraction of clients participates in each communication round. This paper targets the cross-device regime, whose two defining constraints shape our design: (i) each client is selected so rarely that any state persisted on the device between rounds becomes stale before it is next used, and (ii) the server cannot afford to maintain per-client state for a population of millions of devices. Both constraints recur throughout the paper. Data heterogeneity, where each client's local distribution $\mathcal{D}_k$ differs markedly from the global mixture, is therefore the rule rather than the exception, commonly modelled via Dirichlet-partitioned or pathological label-shard splits \cite{LiPartitioning}. Under such Non-IID conditions, multiple local gradient steps cause each client to descend toward its own local minimiser, a phenomenon known as \emph{client drift}, which can slow convergence and introduce systematic bias into the global model \cite{Karimireddy2019,Li2019c,Khaled2019}. FedProx introduces a proximal term in the client objective to penalise local models that deviate from the global model, thereby reducing drift \cite{Li2018}. SCAFFOLD uses control variates at both the server and clients to correct the client update direction and reduce client-drift \cite{Karimireddy2019}. Other approaches such as ADACOMM \cite{Wang2018}, FedBug \cite{Kao2023}, AdaBest \cite{Varno2022}, and Def-KT \cite{Li2020c} also target client-drift directly, while FedNova \cite{Wang2020} tackles the problem by correcting objective inconsistency across clients. Despite their effectiveness, none of these methods exploit the local parameter trajectory to learn a client-specific denoising prior. Moreover, those that persist per-client state on the device (e.g., SCAFFOLD's control variates) degrade under cross-device participation patterns, where clients are revisited too rarely for that state to remain current.


A second, and often more dominant, source of error arises from \emph{partial client participation}: when only a fraction $|\mathcal{S}^{(t)}| \ll N$ of clients contribute to round $t$, the aggregated gradient is a noisy, biased estimator of the true full-participation gradient \cite{Jhunjhunwala2022}. This error has two components: a \emph{sampling} component (which clients happen to be absent in a given round) and a \emph{dispersion} component (how strongly the sampled clients' updates disagree with one another). Because the updates of unsampled clients are simply unavailable, only the server can compensate for the sampling component; the dispersion component, however, can be attacked on-device, by making every client's update less noisy and less drifted before it is sent. Arbitrary device unavailability which creates partial-participation variance has been addressed through server-side \emph{update caching}: MIFA \cite{Gu2021} stores each client's most recent pseudo-gradient on the server and injects these stored updates for absent clients during aggregation, following a SAG-like \cite{Schmidt2014} Variance Reduction (VR) scheme. FedVARP \cite{Jhunjhunwala2022} advances this with a SAGA-like \cite{Defazio2014} estimator; its ClusterFedVARP variant groups clients into clusters to reduce memory, at the cost of cluster-assignment errors and inter-cluster drift. Another recent pre-print, FedAdaVR \cite{fedadavr}, further improves convergence with adaptive variance reduction. However, all three methods (MIFA, FedVARP, and FedAdaVR) require the server to store one full model vector per client, incurring $\mathcal{O}(Nd)$ server-side memory, where $N$ is the total number of clients and $d$ is the model size. At cross-device scale (e.g., $N = 10^6$ mobile devices, modern networks with $d \sim 10^7$ parameters), this storage requirement is entirely impractical: a single 32-bit floating-point model of $10^7$ parameters occupies 40\,MB, so caching even $10^3$ client models demands 40\,GB of server memory.

To address this, we propose \textsc{FedOAED}, a \emph{client-side} mechanism that mitigates client drift at its source and, in doing so, attenuates the dispersion component of partial-participation error, all without incurring any additional server-side memory overhead. Unlike approaches that modify the server aggregation rule \cite{Gu2021,Jhunjhunwala2022,fedadavr}, \textsc{FedOAED} operates entirely on-device. It extracts a denoising prior directly from the client's local optimisation trajectory, while the server only needs to store the global model $\mathbf{w}^{(t)} \in \mathbb{R}^d$ as in standard FedAvg. Our key observation is that a client's own optimisation trajectory, a sequence of parameter snapshots that standard FL simply discards, carries enough structure to learn a client-specific denoising prior on-device, with no server involvement and no reliance on state surviving between rounds. We summarise our main contributions as follows:

\begin{itemize}
        \item We propose \textsc{FedOAED}, in which each client trains an autoencoder on a sliding window of loss-annotated, flattened-parameter snapshots collected during local Stochastic Gradient Descent (SGD). A \emph{loss-weighted} training objective biases the autoencoder toward low-loss parameter regions, resolving the fundamental tension between reconstruction fidelity and gradient-descent loss minimisation. In addition, we introduce three interlocking safety mechanisms to \textsc{FedOAED}, adaptive confidence scaling, a loss-aware rollback gate, and post-refinement gradient recovery steps, that together ensure the autoencoder refinement never degrades the local task loss beyond a tunable tolerance $\rho$.

        \item We show that competitive accuracy under partial participation does not require any server-side client state: \textsc{FedOAED} keeps total server storage at $\mathcal{O}(d)$, the global model alone, yet matches or exceeds, in our experiments, methods such as MIFA and FedVARP, which each require $\mathcal{O}(Nd)$ server memory. Because it also keeps no state on the device across rounds, \textsc{FedOAED} remains practical at cross-device scale, where both server-side and client-side persistence break down.

        \item We establish convergence guarantees for \textsc{FedOAED} in the smooth non-convex setting with partial participation (Section~\ref{sec:theory}). We show that the gated, confidence-scaled refinement is a norm-bounded perturbation whose magnitude is at most $\lambda\tau B_w/4$ per refinement, a factor-of-four improvement contributed precisely by the adaptive confidence scaling, and prove that \textsc{FedOAED} retains FedAvg's $\mathcal{O}\!\bigl(1/\sqrt{M\tilde{K}T}\bigr)$ rate to stationarity (over $T$ rounds with $M$ participating clients and $\tilde{K}$ local steps per round) up to an explicit perturbation neighbourhood, which vanishes under an annealed mixing schedule. This guarantee establishes stability (the refinement cannot destabilise FedAvg), whereas the accuracy gains are demonstrated empirically.

    \item We conduct extensive experiments across multiple Non-IID severity levels on standard benchmarks, demonstrating that \textsc{FedOAED} consistently matches or outperforms state-of-the-art baselines with particularly large gains under extreme heterogeneity.
\end{itemize}

To the best of our knowledge, \textsc{FedOAED} is the first FL algorithm that trains an autoencoder on snapshots of a client's recent parameter trajectory and convexly mixes the reconstruction with the client's current parameters, weighted by a loss-proportional confidence score. Autoencoders have appeared in FL before, for communication compression, anomaly detection, and collaborative representation learning, but never to denoise the local parameter state itself; we defer this discussion to Section~\ref{sec:related}. Crucially, \textsc{FedOAED} achieves this without introducing any server-side storage overhead.

\section{Related Work}\label{sec:related}
 
\paragraph{Federated optimisation and client-drift correction.}
A large body of work addresses client drift arising from multiple local updates under Non-IID data. Regularisation-based methods, for example, FedProx \cite{Li2018}, and FedDyn \cite{Acar2021}, add penalty terms or dynamic bias estimators to the local objective that anchor each client's trajectory to the global model. Control-variate methods like SCAFFOLD \cite{Karimireddy2019}, and Mime \cite{Karimireddy2020} correct the local gradient direction using server-maintained or client-maintained variates. Normalisation-based methods such as FedNova \cite{Wang2020} remove objective inconsistency by rescaling pseudo-gradients before aggregation. MOON \cite{Li2021} takes a contrastive-learning perspective, pulling each client's representation toward the global model's and away from its previous local model's in feature space. Despite their diversity, all of these methods operate on the raw gradient or parameter delta and leave the local parameter trajectory unexploited; none learns a reconstruction prior from that trajectory to actively denoise the client update, which is the core novelty of \textsc{FedOAED}. 
 
\paragraph{Partial client participation and server-side variance reduction.}
Partial participation introduces a sampling-induced variance that is distinct from, and often more damaging than, client drift \cite{Jhunjhunwala2022}. MIFA \cite{Gu2021}, FedVARP \cite{Jhunjhunwala2022}, and recently proposed FedAdaVR \cite{fedadavr} all reduce this variance through server-side update caching, storing a recent pseudo-gradient per client and injecting it for absent clients during aggregation (following SAG- \cite{Schmidt2014} or SAGA-like \cite{Defazio2014} VR principles). As argued in Section~\ref{sec:intro}, this family of methods incurs $\mathcal{O}(Nd)$ server-side memory, which is impractical at massive cross-device scale. Alternative approaches to client unavailability or partial participation, such as compensated aggregation \cite{Su2023}, diverse client sampling \cite{Wang2022}, and heterogeneity-aware aggregation \cite{Mansour2023}, mitigate variance through client selection or aggregation policy, without storing full model copies, but also without any client-side trajectory-based denoising. \textsc{FedOAED} is orthogonal to both families: caching methods compensate for the \emph{sampling} component of participation error, whereas \textsc{FedOAED} shrinks its \emph{dispersion} component by improving the quality of each client's pseudo-gradient before it reaches the server, leaving the server aggregation rule and its memory footprint unchanged; the two are, in principle, combinable.

\paragraph{Autoencoders in federated learning.}
The autoencoder (AE) \cite{Rumelhart1986} is a well-established architecture for unsupervised representation learning: an encoder maps input $x$ to a latent code $z$, and a decoder reconstructs $\hat{x}$ from $z$, trained by minimising a reconstruction loss. In federated learning, autoencoders have been deployed in at least three distinct roles: (i)~\emph{communication compression}---clients encode model updates into a compact latent vector transmitted in lieu of the full parameter delta \cite{Chandar2021}; (ii)~\emph{anomaly and poisoning detection}---a server-side AE trained on benign updates flags outlier contributions \cite{Alonso2024}; and (iii)~\emph{collaborative representation learning}---AE encoder/decoder parameters are aggregated across clients to yield a shared embedding model \cite{Li2025}. None of these roles involves using an AE to reconstruct and denoise the \emph{local parameter state itself} during client training. \textsc{FedOAED} is, to our knowledge, the first method to do so: the AE is trained entirely on-device on the client's own SGD trajectory and its reconstruction is mixed back into local parameters as a denoising step, with no additional server memory cost.
 
\paragraph{Denoising and regularisation in local optimisation.}
Several works aim to stabilise the local SGD step. While momentum-based methods attenuate stochastic gradient noise, and gradient clipping prevents divergence in sharp regions of the loss landscape \cite{Zhang2020}, neither explicitly leverages the historical structure of the parameter trajectory to denoise the update itself. Personalised FL methods such as Per-FedAvg \cite{Fallah2020} and pFedMe \cite{Dinh2020} learn client-specific models or initialisations, but they fundamentally alter the local objective, requiring computationally heavy meta-learning updates, second-order derivative estimations, or dual-model storage on the client device. \textsc{FedOAED} instead learns a \emph{data-free, trajectory-based} denoising prior purely from the sequence of parameter snapshots produced by standard SGD, with no modification to the optimiser, no second-order information, and no additional storage of client-specific states at the server.

\section{Preliminaries}\label{sec:prelim}

We briefly introduce notation used throughout the paper. In the classical supervised learning paradigm, a model is trained on a centralised collection of labelled data samples by minimising a chosen loss function with respect to the model parameters. In federated learning, training is distributed across $N$ clients, each holding a private local dataset $\mathcal{D}_k = \{(x_i, y_i)\}_{i=1}^{n_k}$ without sharing raw data. The objective is to learn a common model $\mathbf{w} \in \mathbb{R}^d$ by jointly minimising the global empirical loss:

{\small
\begin{equation}\label{eq:global_obj}
    \begin{aligned}
    \min_{\mathbf{w}}\; F(\mathbf{w}) = \sum_{k=1}^{N} p_k\,F_k(\mathbf{w}),
    \qquad\\
    F_k(\mathbf{w}) = \frac{1}{n_k}
        \sum_{(x_i,y_i)\in\mathcal{D}_k} \ell\!\bigl(\mathbf{w};\,x_i,y_i\bigr),
    \end{aligned}
\end{equation}
}

\noindent where $p_k = n_k \big/ \sum_{j=1}^{N} n_j$ is client $k$'s data fraction and $\ell(\cdot)$ is the per-sample loss. Data heterogeneity (Non-IID) arises when $F_k \neq F_{k'}$ for $k \neq k'$, which is the common case in cross-device deployments \cite{Zhao2018}. Due to communication limitations and privacy concerns, clients run $K$ local stochastic gradient steps on $F_k$ and transmit only the resulting pseudo-gradient, not raw data to the server. The server aggregates these pseudo-gradients to produce a refined global model, storing only $\mathbf{w}^{(t)} \in \mathbb{R}^d$ at any given round.

\begin{algorithm}
    \caption{\textsc{DeviceUpdate}\((i,\,\mathbf{w}^{(t)},\,\eta_c)\)}
    \label{DeviceUpdateAlgorithm}
    {\small
    \begin{algorithmic}[1]
        \State $\mathbf{w}_i^{(0)} \gets \mathbf{w}^{(t)}$
        \For{local step $k = 0, 1, \ldots, K - 1$}
            \State Compute stochastic gradient
                   $\nabla f_i\!\left(\mathbf{w}_i^{(k)}\right)$
            \State $\mathbf{w}_i^{(k+1)} \gets
                   \mathbf{w}_i^{(k)} - \eta_c\,
                   \nabla f_i\!\left(\mathbf{w}_i^{(k)}\right)$
        \EndFor
        \State \Return
               $\dfrac{1}{\eta_c}\!\left(\mathbf{w}^{(t)} - \mathbf{w}_i^{(K)}\right)$
    \end{algorithmic}
    }
\end{algorithm}

Algorithm~\ref{DeviceUpdateAlgorithm} formalises the standard client update. At round $t$, the server samples a subset $\mathcal{S}^{(t)}$ of clients and broadcasts $\mathbf{w}^{(t)}$. Each selected client runs \textsc{DeviceUpdate} (Algorithm~\ref{DeviceUpdateAlgorithm}), producing the pseudo-gradient

{\small
\begin{equation}\label{eq:pseudograd}
    \mathbf{g}_i^{(t)} =
        \frac{1}{\eta_c}\!\left(\mathbf{w}^{(t)} - \mathbf{w}_i^{(K)}\right),
\end{equation}
}

\noindent where $\eta_c$ is the client learning rate. The server aggregates pseudo-gradients from all participating clients to produce the next global model:

{\small
\begin{equation}\label{FedAvgUpdateEquation}
    \mathbf{w}^{(t+1)} = \mathbf{w}^{(t)} -
        \eta_c \sum_{i \in \mathcal{S}^{(t)}} p_i\,\mathbf{g}_i^{(t)}.
\end{equation}
}

\section{\textsc{FedOAED} Algorithm}
\label{sec:ProposedAlgorithm}

The standard FedAvg procedure is undermined by two compounding error sources. \emph{Client drift}: multiple local steps push $\mathbf{w}_i^{(K)}$ toward the local minimiser of $F_k$ rather than the global minimiser of $F$, introducing systematic bias \cite{Karimireddy2019}. \emph{Participation noise}: when $|\mathcal{S}^{(t)}| \ll N$, the weighted average in Eq.~\eqref{FedAvgUpdateEquation} is a high-variance estimator of the true gradient, slowing convergence under heterogeneous settings \cite{Jhunjhunwala2022}. To address these challenges, specifically, the exacerbation of client drift under limited client availability, we propose \textsc{FedOAED}. Existing remedies for participation noise (e.g., MIFA, FedVARP, FedAdaVR) target the sampling variance by caching client updates on the server, incurring a prohibitive $\mathcal{O}(Nd)$ storage cost that scales poorly with client population $N$ and model size $d$. In contrast, \textsc{FedOAED} operates entirely on the client side; by actively denoising the local trajectory, it mitigates client drift at its source and attenuates the dispersion component of participation noise, all while maintaining a strict $\mathcal{O}(d)$ server storage requirement.

Our approach is inspired by autoencoder-based denoising techniques. While typical neural networks are built for prediction, autoencoder denoisers are designed for representation learning and noise suppression, making them naturally suited to stabilise federated optimisation by filtering out client-specific noise. \textsc{FedOAED} augments the standard FedAvg client update with an on-device autoencoder ($\mathcal{A}_i$). During each communication round, each participating client instantiates an autoencoder $\mathcal{A}_i$, which is trained solely on the client's local parameter trajectory. The server executes standard federated aggregation by broadcasting the global model $\mathbf{w}^{(t)}$, collecting pseudo-gradients, and averaging them.

\paragraph{Local trajectory.} Each participating client runs $E$ local epochs of SGD, i.e.\ $K = E\,b_i$ steps in total. Every $s$ steps it pushes the pair $\bigl(\theta_i^{(k)},\, l_i^{(k)}\bigr)$ onto a \emph{first-in-first-out} (FIFO) buffer $\mathcal{B}$ of capacity $Q$, where $\theta_i^{(k)} = \operatorname{flatten}(\mathbf{w}_i^{(k)}) \in \mathbb{R}^{d}$ concatenates the trainable parameters (buffers such as batch-normalisation running statistics are excluded) and $l_i^{(k)} = f_i(\mathbf{w}_i^{(k-1)}; \xi_{i,k})$ is the mini-batch loss reused from that step's forward pass, so snapshotting adds no gradient or forward computation. Eviction of the oldest entry keeps $\mathcal{B}$ on the most recent segment of the trajectory; without it the buffer saturates early and $\mathcal{A}_i$ is fit to stale, high-loss states. Refinement is applied \emph{in situ} every $E_r$ epochs to the current $\theta_i = \operatorname{flatten}(\mathbf{w}_i)$, using the most recent $W_i = \min(|\mathcal{B}|, W)$ snapshots. To adapt across clients with varying data volumes, $W = \lfloor \gamma\, b_i \rfloor$ is a percentage $\gamma$ of the local batches per epoch $b_i = \lceil n_i/B \rceil$, where $n_i$ is the number of local samples and $B$ the local batch size.

\begin{algorithm}[t]
\caption{\textsc{DeviceUpdate-AE}$(i,\mathbf{w}^{(t)},\eta_c)$}
\label{DeviceUpdateAE}
{\small
\begin{algorithmic}[1]
\State \textbf{Init:} refine coefficient $\lambda$, min snapshots $m$, window percentage $\gamma$, buffer capacity $Q$, snapshot interval $s$, refine cadence $E_r$, AE epochs $e$, chunk fraction $\phi$, rel.\ threshold $\tau$, loss tolerance $\rho$, recovery steps $R$, clip norm $G_{\mathrm{clip}}$, gating batches $b$, temperature $\kappa$, batch size $B$, stability constant $\epsilon_0$; local batches $b_i = \lceil n_i/B \rceil$, model dimension $d$, client optimiser $\textsc{Opt}$ shared by the local and recovery loops; $\operatorname{ch}_C$ zero-pads to a multiple of $C$ and reshapes to rows of width $C$, $\operatorname{un}_d$ inverts and truncates to $d$
\State $W \!\gets\! \lfloor \gamma b_i \rfloor$;\ \ $C \!\gets\! \lfloor \phi d \rfloor$;\ \ 
$\mathbf{w}_i \!\gets\! \mathbf{w}^{(t)}$;\ \ 
$\mathcal{B} \!\gets\! \emptyset$;\ \
       $\mathcal{A}_i \!\gets\! \texttt{None}$
\For{$q = 1,\dots,E$}
  \For{$k = 1,\dots,b_i$}
    \State $l \gets f_i(\mathbf{w}_i;\xi_{i,k})$;\ \
           $\mathbf{w}_i \gets \textsc{Opt}(\mathbf{w}_i, \nabla l)$
           \Comment{unclipped}
    \State \textbf{if} $k \bmod s = 0$ \textbf{then} push
           $(\operatorname{flatten}(\mathbf{w}_i), l)$ onto $\mathcal{B}$,
           evicting oldest while $|\mathcal{B}| > Q$
  \EndFor
  \State \textbf{if} $q \bmod E_r \neq 0$ \textbf{or} $|\mathcal{B}| < m$
         \textbf{then continue}
  \State $(\Theta, L) \gets$ last $W$ of $\mathcal{B}$ if $0 < W < |\mathcal{B}|$,
         else all of $\mathcal{B}$;\ \ $N \gets |\Theta|$
  \State $\zeta \gets (L - \bar{L})/(\operatorname{std}^{\text{pop}}\!(L) + \epsilon_0)$
         if $N > 1$ else $0$;\ \ $\omega \gets \operatorname{softmax}(-\zeta/\kappa)$
  \State $\mathbf{m} \gets \operatorname{mean}_j \Theta_j$;\ \
         $\mathbf{s} \gets \max(\operatorname{std}^{\text{unb}}_j \Theta_j, \epsilon_0)$
         \Comment{elem.-wise}
  \State \textbf{if} $\mathcal{A}_i = \texttt{None}$ \textbf{then} build
         $\mathcal{A}_i\!:\!\mathbb{R}^C\!\to\!\mathbb{R}^C$, $e' \!\gets\! e$
         \textbf{else} $e' \!\gets\! 1$
  \State train $\mathcal{A}_i$ for $e'$ epochs on rows of
         $\operatorname{ch}_C((\Theta - \mathbf{m}) \oslash \mathbf{s})$, minimising
         $\omega$-weighted MSE renormalised per mini-batch
  \State $\widehat{\mathcal{D}}_i \gets$ first $\min(b, b_i)$ batches;\ \
         $\theta_i \gets \operatorname{flatten}(\mathbf{w}_i)$;\ \
         $l_{\mathrm{pre}} \gets \widehat{F}_i(\mathbf{w}_i; \widehat{\mathcal{D}}_i)$
  \State $\widehat{\theta}_i \gets \mathbf{s} \odot \operatorname{un}_d\!\bigl(
         \mathcal{A}_i(\operatorname{ch}_C((\theta_i - \mathbf{m}) \oslash \mathbf{s}))\bigr) + \mathbf{m}$
  \State $r_i \gets \|\theta_i - \widehat{\theta}_i\| / \|\theta_i\|$;\ \
         \textbf{if} $r_i > \tau$ \textbf{then continue}
  \State $c \gets \max(0, 1 - r_i/\tau)$;\ \
         $\widetilde{\theta}_i \gets (1-c\lambda)\theta_i + c\lambda\widehat{\theta}_i$;\ \
         $\mathbf{w}_i \gets \operatorname{unflatten}(\widetilde{\theta}_i)$
  \If{$\widehat{F}_i(\mathbf{w}_i; \widehat{\mathcal{D}}_i) >
       (1+\rho)\, l_{\mathrm{pre}}$}
    \State $\mathbf{w}_i \gets \operatorname{unflatten}(\theta_i)$
           \Comment{rollback}
  \Else
    \State \textbf{for} $j = 1,\dots,R$:\ \
           $\mathbf{w}_i \gets \textsc{Opt}(\mathbf{w}_i,
           \operatorname{clip}_{G_{\mathrm{clip}}}(\nabla f_i(\mathbf{w}_i; \widetilde{\xi}_j)))$
  \EndIf
\EndFor
\State \Return $\frac{1}{\eta_c}(\mathbf{w}^{(t)} - \mathbf{w}_i)$
\end{algorithmic}}
\end{algorithm}

\paragraph{Chunk-level denoiser.} A denoiser of input width $d$ would itself have $\mathcal{O}(dh)$ parameters, which is obsolete on-device. \textsc{FedOAED} therefore shares one autoencoder $\mathcal{A}_i : \mathbb{R}^{C} \to \mathbb{R}^{C}$ across fixed-width chunks, $C = \lfloor \phi d \rfloor$: vectors are zero-padded to a multiple of $C$ and split into $P$ chunks ($\operatorname{ch}_C$), reconstructed in a single batched pass, then concatenated and truncated back to $d$ ($\operatorname{un}_d$). The autoencoder thus has $\mathcal{O}(Ch)$ parameters regardless of $d$, and each snapshot yields $P$ training examples. Under the short local schedules typical of cross-device FL, weight sharing also expands the effective training set from $W_i$ snapshots to $\mathcal{O}(W_i/\phi)$ chunks, making the denoiser trainable from a small window. Both training and inference use coordinates standardised by the element-wise window statistics $\mathbf{m}_i, \mathbf{s}_i$ of the $W_i$ snapshots:

{\small
    \begin{equation}
        \label{eq:recon}
        \widehat{\theta}_i = \mathbf{s}_i \odot \operatorname{un}_d\!\Bigl(
        \mathcal{A}_i\bigl(\operatorname{ch}_C\bigl((\theta_i - \mathbf{m}_i)
        \oslash \mathbf{s}_i\bigr)\bigr)\Bigr) + \mathbf{m}_i .
    \end{equation}
}

This is the denoising mechanism: volatile coordinates are down-weighted by $\mathbf{s}_i$, and the bottleneck cannot reproduce the residual separating $\theta_i$ from the low-loss centre of the window, so $\widehat{\theta}_i$ acts as shrinkage toward that centre rather than as extrapolation. Training is \emph{loss-weighted}: window losses are standardised, $z_j = (l_j - \bar{l})/\operatorname{std}(l)$, and mapped to $\omega_j \propto \exp(-z_j/\kappa)$, so lower-loss snapshots dominate; a snapshot's weight is inherited by all $P$ of its chunks and renormalised within each chunk mini-batch. The bounded softmax avoids the weight collapse of inverse-loss weighting $\omega_j \propto 1/(l_j + \epsilon)$ when window losses span orders of magnitude. The temperature $\kappa$ interpolates between sharply preferring the single best snapshot ($\kappa \to 0$) and uniform weighting ($\kappa \to \infty$); we use $\kappa = 1$.

\paragraph{Confidence gate and rollback.} Two safeguards prevent destructive updates. First, if the relative reconstruction error $r_i = \|\theta_i - \widehat{\theta}_i\| / \|\theta_i\|$ exceeds a threshold $\tau$ the refinement is skipped; otherwise it is scaled by the confidence $c = 1 - r_i/\tau \in [0,1]$, which decays linearly in $r_i$:
{\small
    \begin{equation}
        \label{eq:convex_combination}
        \widetilde{\theta}_i = (1 - c\lambda)\,\theta_i + c\lambda\,\widehat{\theta}_i .
    \end{equation}
}

Second, a \emph{loss-aware rollback gate} compares $l_{\mathrm{pre}} = \widehat{F}_i(\mathbf{w}_i; \widehat{\mathcal{D}}_i)$ and $l_{\mathrm{post}} = \widehat{F}_i(\widetilde{\mathbf{w}}_i; \widehat{\mathcal{D}}_i)$, where $\widehat{F}_i$ averages the loss over a cached set $\widehat{\mathcal{D}}_i$ of $b$ mini-batches with the network in inference mode. Reusing \emph{identical} batches makes the gate deterministic; since $\lambda \ll 1$ the refinement's effect is small by design and would otherwise be confounded with mini-batch noise, causing the gate to reject beneficial updates and accept harmful ones at random. If $l_{\mathrm{post}} > (1+\rho)\, l_{\mathrm{pre}}$ the update is discarded; otherwise the client takes $R$ recovery steps, with gradients clipped to norm $G_{\mathrm{clip}}$, to re-anchor the denoised parameters in the local loss minimum, then uploads the pseudo-gradient:

{\small
    \begin{equation}
    \label{eq:pseudo_gradient}
    \mathbf{g}_i^{(t)} = \frac{1}{\eta_c} \left( \mathbf{w}^{(t)} - \mathbf{w}_i \right)
    \end{equation}
}

The convergence analysis of Section~\ref{sec:theory} treats the variant in which the $R$ recovery steps execute irrespective of the gate outcome; as ordinary SGD steps, this only makes the local step count deterministic without altering the method.

Finally, the server aggregates the updates using standard FedAvg, as illustrated in Equation \ref{FedAvgUpdateEquation} and Algorithm \ref{FedOAEDAlgorithm}.

\begin{algorithm}[t]
    \caption{\textsc{FedOAED}}
    \label{FedOAEDAlgorithm}
    {\small
    \begin{algorithmic}[1]
        \State \textbf{Initialisation:} Set initial model parameters $\mathbf{w}^{(0)}$,
               client learning rate $\eta_c$, number of rounds $T$
        
        \For{$t = 0, 1, \ldots, T-1$}
            \State Send $\mathbf{w}^{(t)}$ to all active devices $i \in \mathcal{S}^{(t)}$
            \State \textit{// Client Side}
            \For{$i \in \mathcal{S}^{(t)}$}
                \State $\mathbf{g}_i^{(t)} \gets
                        \textsc{DeviceUpdate-AE}(i,\,\mathbf{w}^{(t)},\,\eta_c)$
            \EndFor
            \State \textit{// Server Side}
            \State $\mathbf{w}^{(t+1)} =
                    \mathbf{w}^{(t)} - \eta_c \sum_{i \in \mathcal{S}^{(t)}} p_i\,\mathbf{g}_i^{(t)}$
        \EndFor
    \end{algorithmic}
    }
\end{algorithm}

\paragraph{Persistent autoencoder.} Within a round, $\mathcal{A}_i$ is built once at the first refinement cycle (trained for $e$ epochs) and thereafter \emph{incrementally fine-tuned} for a single epoch per cycle on the current window, rather than re-initialised from scratch; the AE optimiser state and the statistics $(\mathbf{m}_i, \mathbf{s}_i)$ are refreshed each cycle, as is the gating set $\widehat{\mathcal{D}}_i$. Across rounds, all \textsc{FedOAED} state ($\mathcal{B}$, $\mathcal{A}_i$, $\mathbf{m}_i$, $\mathbf{s}_i$) is discarded: clients are \emph{stateless between rounds}. This matters under $1$--$2\%$ participation, where stateful client-side schemes (e.g., SCAFFOLD's client control variates) revisit each client so rarely that their persisted state is severely outdated by the next visit.

\paragraph{Complexity.} Let $h$ denote the AE hidden width. The AE operates on chunks of width $C = \phi d$ and adds $\approx 2\phi h d$ parameters on-device (the two $C \times h$ projections dominate). Since each snapshot yields $P = 1/\phi$ chunks and each chunk costs $\mathcal{O}(Ch)$, the chunk count and width cancel: one AE training epoch over a window of $W$ snapshots costs $\mathcal{O}(Whd)$ FLOPs and a reconstruction $\mathcal{O}(hd)$, so chunking reduces AE \emph{parameters} by a factor $\phi$ at unchanged compute. The server stores only $\mathbf{w}^{(t)} \in \mathbb{R}^d$ and aggregates in $\mathcal{O}(d)$ per round, identical to FedAvg, whereas MIFA, FedVARP and FedAdaVR require $\mathcal{O}(Nd)$ server memory that scales with the client population. \textsc{FedOAED} instead places this cost on the $|\mathcal{S}^{(t)}|$ participating clients, where it is transient and parallel: the trajectory buffer, $\mathcal{O}(\min(Q, E b_i / s)\, d)$, dominates on-device memory and is released together with $\mathcal{A}_i$ at the end of the round. In our experiments the buffer holds $15$--$30$ snapshots, so the capacity $Q$ never binds; the term scales with $d$, and for larger backbones is controlled through $\gamma$ and $s$. A local update takes roughly $3\times$ the wall-clock time of FedAvg, trading client-side compute for $\mathcal{O}(d)$ server memory; in cross-device FL, communication latency and server scalability are typically the primary bottlenecks, and client overhead can be reduced via $\gamma$ and $E_r$ without modifying the server.

Section~\ref{sec:theory} shows that the safety gates make this refinement a norm-bounded perturbation of FedAvg, preserving its convergence rate.

\section{Convergence Analysis}\label{sec:theory}

We now show that \textsc{FedOAED}'s three safeguards (the relative-norm gate, adaptive confidence scaling, and the rollback gate) make the autoencoder refinement a \emph{norm-bounded perturbation} of FedAvg, and that \textsc{FedOAED} therefore inherits FedAvg's convergence rate to stationarity up to an explicitly controllable neighbourhood. All proofs are deferred to Appendix~\ref{app:proofs}.

\subsection{Setting and Assumptions}\label{sec:theory_setting}
 
We analyse the equal-weight instantiation of Eq.~\eqref{FedAvgUpdateEquation}: in each round the server selects a subset $|\mathcal{S}^{(t)}| = M$ of the total $N$ clients uniformly at random without replacement (we follow the notation of \cite{Jhunjhunwala2022}), sets $p_i = 1/M$, and aggregates $\mathbf{w}^{(t+1)} = \frac{1}{M}\sum_{i\in\mathcal{S}^{(t)}} \mathbf{w}_i^{\mathrm{fin}}$, where $\mathbf{w}_i^{\mathrm{fin}}$ is client $i$'s final local model. Each client runs $\tilde{K} \coloneqq K + JR$ local SGD steps in total across $J = \lfloor E/E_r \rfloor$ refinement cycles per round ($K$ regular steps plus $R$ recovery steps per cycle; in the analysed variant the $R$ recovery steps execute irrespective of the gate outcome, so the local step count is deterministic---a rejected refinement simply contributes $R$ ordinary SGD steps); since \textsc{FedOAED} has no server optimiser, the aggregated pseudo-step $\eta_c\tilde{K}$ plays the role of the effective step $\tilde{\eta}_s = \eta_s\eta_c K$ in server methods \cite{Jhunjhunwala2022}. We write $\Delta \coloneqq F(\mathbf{w}^{(0)}) - F^\ast$ with $F^\ast \coloneqq \inf_{\mathbf{w}} F(\mathbf{w}) > -\infty$.

\begin{assumption}[Lipschitz gradient]\label{ass:smooth}
Each per-sample loss $f_i(\cdot\,;\xi)$ is differentiable and there exists a constant $L$ such that $\|\nabla f_i(\mathbf{u};\xi) - \nabla f_i(\mathbf{v};\xi)\| \le L\|\mathbf{u}-\mathbf{v}\|$ for all $\mathbf{u},\mathbf{v}$, almost surely; consequently each $F_i$ and $F$ are $L$-smooth.
\end{assumption}
 
\begin{assumption}[Bounded local variance]\label{ass:var}
The stochastic gradient at each client is an unbiased estimator of the local gradient with bounded variance: $\mathbb{E}_{\xi}\bigl[\nabla f_i(\mathbf{w};\xi)\bigr] = \nabla F_i(\mathbf{w})$ and $\mathbb{E}_{\xi}\bigl\|\nabla f_i(\mathbf{w};\xi) - \nabla F_i(\mathbf{w})\bigr\|^2 \le \sigma^2$ for all $i, \mathbf{w}$.
\end{assumption}
 
\begin{assumption}[Bounded gradients]\label{ass:clip}
$\|\nabla f_i(\mathbf{w};\xi)\| \le G$ almost surely. (If the implementation additionally applies gradient clipping with a threshold $G_{\mathrm{clip}} \ge G$, the clip operator is the identity along the trajectory and can be ignored throughout the analysis.) By Jensen's inequality, $\|\nabla F_i(\mathbf{w})\| \le G$ and $\|\nabla F(\mathbf{w})\| \le G$; in particular this subsumes the usual bounded-heterogeneity assumption, since $\frac1N\sum_i\|\nabla F_i(\mathbf{w})-\nabla F(\mathbf{w})\|^2 \le 4G^2$.
\end{assumption}

\begin{assumption}[Bounded iterates]\label{ass:bounded}
All local iterates remain in a ball: $\|\mathbf{w}_i^{(k)}\| \le B_w$ along the trajectory.
\end{assumption}
 
Assumptions~\ref{ass:smooth} and~\ref{ass:clip} are standard in the nonconvex optimisation literature, and Assumption~\ref{ass:var} is widely adopted in federated optimisation \cite{Karimireddy2019,Gu2021,Jhunjhunwala2022}. Together they are the sample-wise counterparts of the assumptions of FedVARP \cite{Jhunjhunwala2022} (Lipschitz gradient, bounded variance, bounded gradients), \emph{minus} its global-variance parameter $\sigma_g$, which our analysis does not require: Assumption~\ref{ass:clip} already gives $\sigma_g \le 2G$, and trading $\sigma_g$ for $G$ keeps every constant small and elementary. The sample-wise form is what gradient clipping (when used) and the coupling argument of Lemma~\ref{lem:prop} act on; clipped-FL techniques \cite{Zhang2020} could relax the inactive-clipping requirement. Assumption~\ref{ass:bounded} needed only to convert the \emph{relative}-norm gate into an \emph{absolute} budget, which effectively concerns the global iterates and can be enforced exactly by projection (Remark~\ref{rem:A5}).

\subsection{Main Result}
 
\begin{theorem}[Non-convex convergence of \textsc{FedOAED}]\label{thm:main}
Suppose Assumptions~\ref{ass:smooth}--\ref{ass:bounded} hold. In each round of \textsc{FedOAED}, the server selects $|\mathcal{S}^{(t)}| = M$ of the total $N$ clients uniformly at random, each conducting $\tilde{K}$ local SGD steps with $J$ gated refinements. If the client learning rate satisfies $\eta_c \le \frac{1}{8 L \tilde{K}}$, then the iterates $\{\mathbf{w}^{(t)}\}$ of \textsc{FedOAED} satisfy
{\small
\begin{equation}\label{eq:main_bound}
\begin{aligned}
\frac{1}{T}\sum_{t=0}^{T-1} \mathbb{E}\bigl\|\nabla F(\mathbf{w}^{(t)})\bigr\|^2
\;\le\;& \frac{8\,\Delta}{3\,\eta_c \tilde{K} T}
\;+\; \frac{5\,\eta_c L \sigma^2}{M}+\; \frac{18\,\eta_c L \tilde{K}\, G^2}{M}
\;+\; \frac{\eta_c^2 L^2 \tilde{K}^2 G^2}{2}\\
&+\; \underbrace{\frac{11\,\bar{\varepsilon}^2}{\eta_c^2 \tilde{K}^2} \;+\; \frac{2 L\,\bar{\varepsilon}^2}{\eta_c \tilde{K}}}_{\textstyle \Phi(\bar{\varepsilon}) \text{: refinement neighbourhood}} .
\end{aligned}
\end{equation}
}
\end{theorem}
 
Setting $\lambda = 0$ (hence $\bar{\varepsilon} = 0$) recovers a FedAvg-type bound, confirming that the analysis is tight in the unperturbed limit; since the minimum is below the average, the same bound holds verbatim for $\min_{0\le t\le T-1}\mathbb{E}\|\nabla F(\mathbf{w}^{(t)})\|^2$, the form reported in \cite{Gu2021,Jhunjhunwala2022}. In the language of the FedVARP error decomposition \cite{Jhunjhunwala2022}, the second term of Eq.~\eqref{eq:main_bound} is the stochastic-gradient error, the third is the partial-participation error, the fourth is the client-drift error, and $\Phi(\bar{\varepsilon})$ is the \emph{refinement error}---the only new term, contributed solely through an additive neighbourhood that the practitioner controls through $\lambda$ and $\tau$. Note that the participation term is bounded here via the worst-case dispersion $4G^2$ supplied by Assumption~\ref{ass:clip}; the mechanism by which \textsc{FedOAED} empirically shrinks the \emph{realised} dispersion of client updates is deliberately not modelled (Remark~\ref{rem:scope}). The $G^2$-dependence of the participation and drift terms is the price of the elementary bounded-gradient analysis; a sharper variant separating the heterogeneity parameter $\sigma_g \le 2G$ (as in \cite{Jhunjhunwala2022}) is possible at the cost of a substantially longer drift recursion, and we deliberately prefer the simple constants above.

\begin{corollary}\label{cor:rate}
Suppose the conditions in Theorem~\ref{thm:main} are satisfied, and choose $\eta_c = \min\bigl\{\frac{1}{8L\tilde{K}},\, \sqrt{\frac{M\Delta}{L\tilde{K}T\sigma^2}}\bigr\}$. Then the first four terms of Eq.~\eqref{eq:main_bound} are $\mathcal{O}\bigl(\sigma\sqrt{L\Delta}/\sqrt{M\tilde{K}T}\bigr) + \mathcal{O}(1/T)$ (treating $\tilde{K}$ and $G$ as constants; the participation term retains the $G^2$ factor discussed above), matching FedAvg under partial participation, with all constants explicit in Appendix~\ref{app:proof_cor}. For the neighbourhood: with constant $\lambda$, \textsc{FedOAED} converges to a stationarity neighbourhood of size $\Phi(\bar{\varepsilon})$, and $\Phi(\bar{\varepsilon}) \le 12\chi^2$ whenever $\bar{\varepsilon} \le \chi\,\eta_c\tilde{K}$ (perturbation no larger than $\chi$ times one aggregated pseudo-step); with annealed mixing $\lambda_t = \lambda_0(t+1)^{-1/4}$, the time-averaged neighbourhood is at most $24\,\chi_0^2/\sqrt{T}$, so \textsc{FedOAED} converges \emph{exactly} to stationarity at the unchanged leading rate.
\end{corollary}

\begin{remark}\label{rem:scope}
Theorem~\ref{thm:main} treats the autoencoder reconstruction as an arbitrary perturbation within the gated budget and shows the safety mechanisms suffice to preserve FedAvg's rate. Analogous perturbation-tolerant analyses are the standard first theoretical step for denoising and compression mechanisms in FL \cite{Jhunjhunwala2022}.
\end{remark}

\section{Experiments}\label{sec:experiments}

\subsection{Experimental Setup}\label{ExperimentalSetup}

All experiments were implemented in PyTorch~\cite{PyTorchPaper} using the Flower framework~\cite{FlowerPaper}. To evaluate across modalities, we utilise F-MNIST~\cite{xiao2017fmnist} and CIFAR-10~\cite{Krizhevsky09cifar10} for image classification under various synthetic non-IID partitions (IID/Non-IID, Dirichlet $\alpha=0.5$, LQ-2, and LQ-3)~\cite{LiPartitioning}, alongside the Shakespeare dataset~\cite{Caldas2018} for a naturally partitioned language modelling task. Model architectures and client distributions are assigned as follows: LeNet-5~\cite{Lecun1998} for F-MNIST (500 total clients, 1\% active per round); ResNet-18~\cite{resnet} with group normalisation~\cite{Hsieh2019} replacing batch normalisation for CIFAR-10 (250 clients, 2\% active); and a GRU-based RNN~\cite{Chung2014} for Shakespeare (904 training and 225 testing clients, $\leq$1\% active). At each round, clients are sampled uniformly at random without replacement. This combination of severe label skew and low participation yields a highly challenging federated learning environment. The FedVARP and MIFA baselines (using a fixed client learning rate) were reproduced independently. Comprehensive experimental details, including exhaustive hyperparameter configurations (batch sizes, learning rates, seeds, and autoencoder dimensions) and PyTorch-style pseudocode for \textsc{FedOAED}’s autoencoders, are provided in Appendices~\ref{Apd:ExpSetup} and~\ref{app:implementation}.

\subsection{Comparison with State-of-the-Art Methods}\label{ExperimentalResultsComparisonSOTA}

\begin{figure*}[!ht]
	\centering
	\includegraphics[scale=0.105]{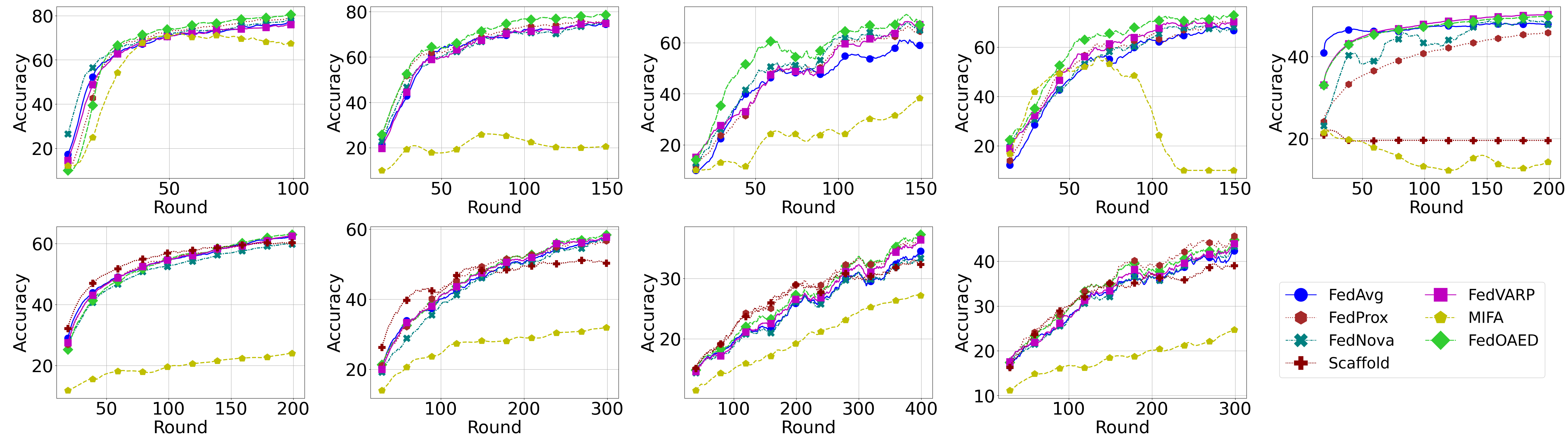}
	\caption{Accuracy comparison across various data partitioning methods. The top row (left to right) illustrates IID/Non-IID, Dirichlet, LQ-2, and LQ-3 splits on the FMNIST dataset, followed by the IdPartitioner on the Shakespeare dataset. The bottom row depicts the corresponding IID/Non-IID, Dirichlet, LQ-2, and LQ-3 configurations on the CIFAR-10 dataset.}
	\label{fig:final_evaluation_accuracy}
\end{figure*}

We evaluate \textsc{FedOAED} against six state-of-the-art baselines: FedAvg~\cite{McMahan2016}, FedProx~\cite{Li2018}, SCAFFOLD~\cite{Karimireddy2019}, FedNova~\cite{Wang2020}, MIFA~\cite{Gu2021}, and FedVARP~\cite{Jhunjhunwala2022}, utilising accuracy as the primary metric visualised in Figure~\ref{fig:final_evaluation_accuracy}. 

Under severe data heterogeneity and limited participation, established VR baselines exhibited instability. While SCAFFOLD initially converged, it failed on the naturally partitioned Shakespeare dataset, plateauing near $20\%$. MIFA achieved suboptimal accuracy and frequently suffered late-stage divergence due to stale gradients, reaching only $23.78\%$ on CIFAR-10 IID-NonIID and collapsing entirely on F-MNIST LQ-3. Conversely, FedAvg and FedNova proved unexpectedly robust, highlighting the difficulty of maintaining stable variance reduction without full client participation.

\textsc{FedOAED} achieved the highest accuracy across all four F-MNIST partitions, reaching $79.56\%$ on IID-NonIID and $78.12\%$ on Dirichlet ($\alpha=0.5$), compared with $75.28\%$ for FedAvg, $75.36\%$ for FedProx, and $75.44\%$ for FedVARP on the Dirichlet split. On CIFAR-10, \textsc{FedOAED} similarly led on IID-NonIID ($62.85\%$ vs.\ $62.30\%$ for FedVARP), Dirichlet ($58.39\%$), and LQ-2 ($37.29\%$). The only exception was CIFAR-10 LQ-3, where \textsc{FedOAED} ($44.79\%$) trailed FedProx ($45.59\%$) while remaining ahead of FedVARP ($43.95\%$) and FedAvg ($42.35\%$).

On the naturally partitioned Shakespeare benchmark, FedVARP reached the highest accuracy (roughly $50\%$), with \textsc{FedOAED} following closely and outperforming FedProx (which plateaued near $45\%$). These results indicate that on-device trajectory denoising mitigates client drift without relying on historical server-side updates. Full numerical results are reported in Appendix~\ref{Apd:ExpResults}.

In all settings, \textsc{FedOAED} avoids both SCAFFOLD's additional communication overhead and the $\mathcal{O}(Nd)$ server memory required by MIFA and FedVARP, since no per-client history is cached. Ablation studies on the rollback gate, loss-weighted sampling, and hyperparameter sensitivity are provided in Appendices~\ref{app:ablation} and~\ref{app:sensitivity_analysis}.

\section{Conclusion and Future Work}\label{sec:conclusion}

This paper proposes \textsc{FedOAED}, a federated learning algorithm that mitigates client drift under limited client availability. By integrating a client-side autoencoder, \textsc{FedOAED} reconstructs denoised updates from recent flattened-parameter snapshots, convexly mixing them into the final client update prior to server aggregation. Empirically, \textsc{FedOAED} substantially improves convergence across diverse datasets and architectures. Theoretically, this gated refinement constitutes a norm-bounded perturbation of FedAvg, retaining its $\mathcal{O}\bigl(1/\sqrt{M\tilde{K}T}\bigr)$ non-convex stationarity rate up to an explicitly controllable neighbourhood that vanishes under an annealed mixing schedule (Section~\ref{sec:theory}).

\textbf{Limitations and Future Work.} \textsc{FedOAED}'s primary trade-off is on-device cost: training the autoencoder makes a local update $\approx 3\times$ slower than FedAvg, and the client transiently holds a trajectory buffer of $\mathcal{O}(W d)$, both released each round (Appendices~\ref{app:compute_profile} and~\ref{app:limitations}). For larger backbones the buffer binds, and is controlled by $\gamma$ and the snapshot interval $s$; our sensitivity analysis shows shorter windows match or exceed the base configuration (Appendix~\ref{app:sensitivity_analysis}). Future work will explore resource-adaptive dynamic autoencoders, a wider range of autoencoder architectures, and bespoke hyperparameter tuning beyond the standard CIFAR-10 and Shakespeare settings.

\bibliography{references_file}
\bibliographystyle{plainnat}


\appendix

\clearpage

\addcontentsline{toc}{section}{Appendix}
\tableofcontents
\newpage

\section*{Appendix Overview}

Appendix~\ref{app:proofs} contains the convergence analysis deferred from Section~\ref{sec:theory}. Appendix~\ref{Apd:ExpSetup} records the datasets, partitioning schemes and federated configuration; Appendix~\ref{app:implementation} the implementation details required to reproduce \textsc{FedOAED} exactly, including the autoencoder architectures and the numerical safeguards omitted from Algorithm~\ref{DeviceUpdateAE} for space. Appendix~\ref{Apd:ExpResults} gives the full results underlying Section~\ref{sec:experiments}, and Appendix~\ref{app:compute_profile} profiles the computational and memory cost of the on-device machinery. Appendices~\ref{app:ablation} and~\ref{app:sensitivity_analysis} establish, respectively, that each novel mechanism contributes to the result and that the method is not delicately dependent on its hyperparameters; Appendix~\ref{app:tuning_guidelines} converts those measurements into tuning guidance. Appendix~\ref{app:limitations} states the limitations of the work in full.


\section{Convergence Analysis for \textsc{FedOAED}}\label{app:proofs}

\subsection{Basic Notation}\label{app:facts}

Fix a round $t$ and drop the superscript: $\mathbf{x} \coloneqq \mathbf{w}^{(t)}$, $\eta \coloneqq \eta_c$. For \emph{every} client $i \in [N]$ (not only sampled ones), define the \emph{virtual unrefined} trajectory
\begin{align*}
\mathbf{y}_{i,0} & = \mathbf{x}, \\
\mathbf{g}_{i,k} & \coloneqq \nabla f_i(\mathbf{y}_{i,k};\xi_{i,k}), \\
\mathbf{y}_{i,k+1} & = \mathbf{y}_{i,k} - \eta\,\mathbf{g}_{i,k},
\quad k = 0,\dots,\tilde{K}-1,
\end{align*}
driven by the same sample sequence $\{\xi_{i,k}\}$ as the actual run. Write $\bar{\mathbf{y}}_i \coloneqq \mathbf{y}_{i,\tilde{K}}$, let $\widehat{\mathbf{w}}_i$ denote client $i$'s actual final model (with refinement), and define
\begin{align*}
\bar{\mathbf{x}}^{+} & \coloneqq
\frac{1}{M}\sum_{i\in\mathcal{S}}\bar{\mathbf{y}}_i, \\
\mathbf{w}^{+} & \coloneqq
\frac{1}{M}\sum_{i\in\mathcal{S}}\widehat{\mathbf{w}}_i, \\
\bar{\mathbf{e}} & \coloneqq
\mathbf{w}^{+} - \bar{\mathbf{x}}^{+}.
\end{align*}
Write also $\bar{\mathbf{g}}_k \coloneqq \frac{1}{M}\sum_{i\in\mathcal{S}} \mathbf{g}_{i,k}$, so that $\bar{\mathbf{x}}^{+} - \mathbf{x} = -\eta\sum_{k=0}^{\tilde{K}-1}\bar{\mathbf{g}}_k$. The sampled client set $\mathcal{S}$ is drawn independently of all $\{\xi_{i,k}\}$ and hence of the virtual trajectories. To make the information structure explicit, let $\mathcal{G}_k \coloneqq \sigma\bigl(\{\xi_{i,j} : i \in [N],\ j \le k-1\}\bigr)$ denote the trajectory filtration. Then every $\mathbf{y}_{i,k}$ is $\mathcal{G}_k$-measurable, the fresh sample $\xi_{i,k}$ is independent of $\mathcal{G}_k$, the samples $\{\xi_{i,k}\}_{i\in[N]}$ at any fixed step $k$ are mutually independent across clients, and $\mathcal{S} \perp \mathcal{G}_\infty$. All virtual trajectories are defined on the same product probability space as the actual run, so defining them for unsampled clients creates no measurability issue.

\subsection{Auxiliary Lemmas}\label{app:aux}

\begin{lemma}[Young's inequality]\label{lem:young}
For any $\mathbf{u},\mathbf{v}\in\mathbb{R}^d$ and $\nu>0$, $2\langle \mathbf{u},\mathbf{v}\rangle \le \nu\,\|\mathbf{u}\|^2 + \nu^{-1}\|\mathbf{v}\|^2$.
\end{lemma}

\begin{lemma}[Jensen's inequality]\label{lem:jensen}
For vectors $\mathbf{a}_1,\dots,\mathbf{a}_n\in\mathbb{R}^d$, $\bigl\|\tfrac1n\sum_{k=1}^{n} \mathbf{a}_k\bigr\|^2 \le \tfrac1n\sum_{k=1}^{n}\|\mathbf{a}_k\|^2$.
\end{lemma}

\begin{lemma}[Sum of squares]\label{lem:sumsq}
For a positive integer $n$ and vectors $\mathbf{a}_1,\dots,\mathbf{a}_n\in\mathbb{R}^d$, $\bigl\|\sum_{k=1}^{n} \mathbf{a}_k\bigr\|^2 \le n\sum_{k=1}^{n} \|\mathbf{a}_k\|^2$.
\end{lemma}

\begin{lemma}[Smoothness descent]\label{lem:descineq}
If $F$ is $L$-smooth, then $F(\mathbf{v}) \le F(\mathbf{u}) + \langle\nabla F(\mathbf{u}),\, \mathbf{v}-\mathbf{u}\rangle + \tfrac{L}{2}\|\mathbf{v}-\mathbf{u}\|^2$ for all $\mathbf{u},\mathbf{v}$.
\end{lemma}

\begin{lemma}[Variance under uniform, without-replacement sampling]\label{lem:sampling}
Let $\bar{a} = \frac1N\sum_{i=1}^{N} a_i$. If $\mathcal{S}$ of size $M$ is sampled uniformly at random without replacement from $[N]$, then $\mathbb{E}\bigl[\tfrac1M\sum_{i\in\mathcal{S}} a_i\bigr] = \bar{a}$ and
\begin{equation*}
\begin{aligned}
\mathbb{E}_{\mathcal{S}}\Bigl\|
\tfrac1M\textstyle\sum_{i\in\mathcal{S}} a_i - \bar{a}
\Bigr\|^2
&= \frac{1}{M}\,\frac{N-M}{N-1}
\cdot\frac1N\sum_{i=1}^{N} \|a_i-\bar{a}\|^2 \\
&\le \frac{1}{M}\cdot\frac1N
\sum_{i=1}^{N}\|a_i-\bar{a}\|^2 .
\end{aligned}
\end{equation*}
\end{lemma}

\begin{proof}
Let $I_i \coloneqq \mathbb{1}\{i \in \mathcal{S}\}$ be the inclusion indicators. Uniform without-replacement sampling of a fixed-size set gives $\sum_i I_i = M$ exactly, $\mathbb{E}[I_i] = M/N$, and $\mathbb{E}[I_i I_j] = \frac{M(M-1)}{N(N-1)}$ for $i \neq j$. The mean claim is immediate: $\mathbb{E}\bigl[\tfrac1M\sum_i I_i a_i\bigr] = \tfrac1M\sum_i \tfrac{M}{N} a_i = \bar{a}$. For the variance, set $\mathbf{b}_i \coloneqq a_i - \bar{a}$, so $\sum_i \mathbf{b}_i = 0$; because $\sum_i I_i = M$ deterministically,

\begin{equation*}
\begin{aligned}
\frac1M\sum_{i\in\mathcal{S}} a_i - \bar{a}
&= \frac1M\sum_{i=1}^{N} I_i\,\mathbf{b}_i, \\
\mathbb{E}\Bigl\|\frac1M\sum_i I_i \mathbf{b}_i\Bigr\|^2
&= \frac{1}{M^2}\Biggl[
\frac{M}{N}\sum_i \|\mathbf{b}_i\|^2 + \frac{M(M-1)}{N(N-1)}
\sum_{i\neq j}\langle \mathbf{b}_i,\mathbf{b}_j\rangle
\Biggr].
\end{aligned}
\end{equation*}
Since $\sum_{i\neq j}\langle \mathbf{b}_i,\mathbf{b}_j\rangle = \bigl\|\sum_i \mathbf{b}_i\bigr\|^2 - \sum_i\|\mathbf{b}_i\|^2 = -\sum_i\|\mathbf{b}_i\|^2$, the bracket equals $\frac{M}{N}\bigl(1 - \frac{M-1}{N-1}\bigr)\sum_i\|\mathbf{b}_i\|^2 = \frac{M}{N}\,\frac{N-M}{N-1}\sum_i\|\mathbf{b}_i\|^2$, and dividing by $M^2$ gives the stated identity; the inequality follows from $\frac{N-M}{N-1} \le 1$.
\end{proof}

\begin{lemma}[Perturbation budget]\label{lem:budget}
Under Assumption~\ref{ass:bounded}, the parameter change $\boldsymbol{\delta}_i \coloneqq \widetilde{\mathbf{w}}_i - \mathbf{w}_i$ induced by one attempted refinement of Algorithm~\ref{DeviceUpdateAE}, whether accepted or rejected, satisfies
\begin{equation}\label{eq:budget}
\|\boldsymbol{\delta}_i\| \;\le\; \frac{\lambda\,\tau}{4}\, B_w \;\eqqcolon\; \varepsilon .
\end{equation}
If adaptive confidence scaling is removed while retaining the same relative-norm gate (i.e., $c \equiv 1$), the corresponding uniform worst-case bound becomes $\lambda\,\tau B_w$, which is four times the bound above.
\end{lemma}

\begin{proof}
If the relative-norm gate rejects ($r > \tau$) or the rollback gate rejects, then $\boldsymbol{\delta}_i = 0$ and \eqref{eq:budget} holds trivially. Otherwise, with $\theta_i = \operatorname{flatten}(\mathbf{w}_i)$, $r = \|\widehat{\theta}_i - \theta_i\|/\|\theta_i\| \le \tau$ and $c = 1 - r/\tau$,
\begin{equation*}
\|\boldsymbol{\delta}_i\| \;=\; c\lambda\,\bigl\|\widehat{\theta}_i - \theta_i\bigr\| \;=\; \lambda\Bigl(1-\frac{r}{\tau}\Bigr)\, r\,\|\theta_i\| .
\end{equation*}

Since $h(r) = r(1 - r/\tau)$ on $[0,\tau]$ is maximised at $r = \tau/2$ with value $h(\tau/2) = \tau/4$, Assumption~\ref{ass:bounded} ($\|\theta_i\| \le B_w$) yields $\|\boldsymbol{\delta}_i\| \le \lambda\tau B_w/4$. Without adaptive scaling ($c \equiv 1$), $\|\boldsymbol{\delta}_i\| = \lambda r\|\theta_i\| \le \lambda\tau B_w$.
\end{proof}

\begin{remark}[On Assumption~\ref{ass:bounded}]\label{rem:A5}
Within a round, Assumption~\ref{ass:clip} already implies $\|\mathbf{w}_i^{(k)} - \mathbf{w}^{(t)}\| \le \eta\tilde{K} G \le G/(8L)$, so Assumption~\ref{ass:bounded} effectively constrains only the global iterates $\{\mathbf{w}^{(t)}\}_{t \le T}$. It can be enforced directly by projecting the aggregated server model onto the ball of radius $B_w$ at the end of each round (since projection onto a convex set is non-expansive, the proof below is unchanged), or via weight decay.
\end{remark}

\begin{lemma}[Propagation]\label{lem:prop}
Under Assumptions~\ref{ass:smooth} and~\ref{ass:clip}, with $\eta \le \frac{1}{8L\tilde{K}}$ and $J$ refinement cycles per round, each satisfying Lemma~\ref{lem:budget}, the following bound holds almost surely:
\begin{equation*}
\begin{aligned}
\bigl\|\widehat{\mathbf{w}}_i - \bar{\mathbf{y}}_i\bigr\|
&\le J\,(1+\eta L)^{\tilde{K}}\,\varepsilon \\
&\le J\,e^{\eta L\tilde{K}}\,\varepsilon \\
&\le 1.14\,J\,\varepsilon
\eqqcolon \bar{\varepsilon}.
\end{aligned}
\end{equation*}
\end{lemma}

\begin{proof}
Throughout, the actual and virtual trajectories each take exactly $\tilde{K} = K + JR$ steps driven by the same sample sequence $\{\xi_{i,k}\}$. Making the $R$ recovery steps unconditional ensures that the two trajectories have equal step counts, so the synchronous coupling below is well defined.
For two sequences driven by the \emph{same} samples, $\mathbf{u}_{k+1} = \mathbf{u}_k - \eta\nabla f_i(\mathbf{u}_k;\xi_k)$ and $\mathbf{v}_{k+1} = \mathbf{v}_k - \eta\nabla f_i(\mathbf{v}_k;\xi_k)$, per-sample $L$-smoothness gives
\begin{equation*}
\begin{aligned}
\|\mathbf{u}_{k+1}-\mathbf{v}_{k+1}\|
&\le \|\mathbf{u}_k-\mathbf{v}_k\| + \eta\bigl\|\nabla f_i(\mathbf{u}_k;\xi_k)
- \nabla f_i(\mathbf{v}_k;\xi_k)\bigr\| \\
&\le (1+\eta L)\,\|\mathbf{u}_k-\mathbf{v}_k\|,
\end{aligned}
\end{equation*}
i.e., each same-sample SGD map is $(1+\eta L)$-Lipschitz. 
Now let $\Gamma_k \coloneqq \|\mathbf{w}_{i,k}^{\mathrm{act}} - \mathbf{y}_{i,k}\|$ denote the gap between the actual and the virtual trajectory after $k$ steps, and let $t_1 < \dots < t_J$ be the (deterministic, epoch-boundary) steps after which the $J$ refinements are injected. Between injections the gap satisfies $\Gamma_{k+1} \le (1+\eta L)\,\Gamma_k$ by the display above; at an injection the actual iterate is additionally displaced by $\boldsymbol{\delta}_i^{(j)}$ with $\|\boldsymbol{\delta}_i^{(j)}\| \le \varepsilon$ pathwise (Lemma~\ref{lem:budget}, covering acceptance, rejection, and rollback alike). The gap therefore obeys the linear recursion
\begin{equation*}
\begin{aligned}
\Gamma_{k+1}
&\le (1+\eta L)\,\Gamma_k + \varepsilon\,\mathbb{1}
\{k+1 \in \{t_1,\dots,t_J\}\}, \quad
\Gamma_0 = 0,
\end{aligned}
\end{equation*}
whose unrolled form is
\begin{equation*}
\Gamma_{\tilde{K}} \;\le\; \sum_{j=1}^{J} (1+\eta L)^{\tilde{K}-t_j}\,\varepsilon
\;\le\; J\,(1+\eta L)^{\tilde{K}}\,\varepsilon .
\end{equation*}
The recursion is linear in the gap, so the $J$ contributions add without cross terms. 

Finally $(1+\eta L)^{\tilde{K}} \le e^{\eta L\tilde{K}} \le e^{1/8} \le 1.14$. Consequently, $\|\bar{\mathbf{e}}\| \le \frac1M\sum_{i\in\mathcal{S}}\|\widehat{\mathbf{w}}_i - \bar{\mathbf{y}}_i\| \le \bar{\varepsilon}$ almost surely.
\end{proof}

\begin{lemma}[Drift]\label{lem:drift}
Let Assumption~\ref{ass:clip} hold. Define $d_k \coloneqq \frac1N\sum_{i=1}^N \mathbb{E}\|\mathbf{y}_{i,k}-\mathbf{x}\|^2$ and $D \coloneqq \sum_{k=0}^{\tilde{K}-1} d_k$. Then $\|\mathbf{y}_{i,k}-\mathbf{x}\| \le \eta\,k\,G$ holds almost surely for every $i$ and every $k \le \tilde{K}$, and consequently
\begin{equation}\label{eq:drift}
d_k \;\le\; \eta^2 k^2 G^2, \qquad D \;\le\; \frac{\eta^2 G^2 \tilde{K}^3}{3}.
\end{equation}
\end{lemma}

\begin{proof}
$\mathbf{y}_{i,k} - \mathbf{x} = -\eta\sum_{j<k}\mathbf{g}_{i,j}$ with each $\|\mathbf{g}_{i,j}\| \le G$ (Assumption~\ref{ass:clip}), so $\|\mathbf{y}_{i,k}-\mathbf{x}\| \le \eta k G$ by the triangle inequality. 

Squaring and averaging gives $d_k \le \eta^2k^2G^2$, and summing over $k=0,\dots,\tilde{K}-1$ yields $D \le \eta^2G^2\sum_{k=0}^{\tilde{K}-1}k^2 \le \eta^2G^2\tilde{K}^3/3$.

\end{proof}

\begin{lemma}[One-round descent]\label{lem:descent}
Let Assumptions~\ref{ass:smooth}--\ref{ass:bounded} hold and $\eta \le \frac{1}{8L\tilde{K}}$. Then for every round $t$,
\begin{equation}\label{eq:one_round}
\begin{aligned}
\mathbb{E}\bigl[F(\mathbf{w}^{(t+1)})\bigr]
&\le \mathbb{E}\bigl[F(\mathbf{w}^{(t)})\bigr] - \frac{3\,\eta\tilde{K}}{8}\,
\mathbb{E}\bigl\|\nabla F(\mathbf{w}^{(t)})\bigr\|^2 \\
&\quad + \frac{8\,\eta^2 L \tilde{K}\sigma^2}{5M}
+ \frac{32\,\eta^2 L \tilde{K}^2 G^2}{5M} + \frac{\eta^3 L^2 G^2 \tilde{K}^3}{6}
+ \frac{4\,\bar{\varepsilon}^2}{\eta\tilde{K}}
+ \frac{L\,\bar{\varepsilon}^2}{2}.
\end{aligned}
\end{equation}
\end{lemma}

\begin{proof}
All expectations are conditional on $\mathbf{x} = \mathbf{w}^{(t)}$; \eqref{eq:one_round} follows by the tower rule.

\textbf{Step 1 (two smoothness expansions).} By Lemma~\ref{lem:prop}, $\mathbf{w}^{+} = \bar{\mathbf{x}}^{+} + \bar{\mathbf{e}}$ with $\|\bar{\mathbf{e}}\| \le \bar{\varepsilon}$ almost surely. Applying Lemma~\ref{lem:descineq} twice---first at $\bar{\mathbf{x}}^{+}$ in the direction $\bar{\mathbf{e}}$, then at $\mathbf{x}$ in the direction $\bar{\mathbf{x}}^{+}-\mathbf{x}$---gives
\begin{equation*}
\begin{aligned}
F(\mathbf{w}^{+})
&\le F(\mathbf{x})
+ \langle \nabla F(\mathbf{x}),\bar{\mathbf{x}}^{+}-\mathbf{x}\rangle + \tfrac{L}{2}\bigl\|\bar{\mathbf{x}}^{+}-\mathbf{x}\bigr\|^2 + \langle \nabla F(\bar{\mathbf{x}}^{+}),\bar{\mathbf{e}}\rangle
+ \tfrac{L}{2}\bar{\varepsilon}^2.
\end{aligned}
\end{equation*}

\textbf{Step 2 (perturbation cross term).} Applying Lemma~\ref{lem:young} with $\nu = \eta\tilde{K}/8$ and using $\|\nabla F(\bar{\mathbf{x}}^{+})\|^2 \le 2\|\nabla F(\mathbf{x})\|^2 + 2L^2\|\bar{\mathbf{x}}^{+}-\mathbf{x}\|^2$ (by $L$-smoothness and Lemma~\ref{lem:sumsq}) gives

\begin{equation*}
\begin{aligned}
\langle \nabla F(\bar{\mathbf{x}}^{+}),\bar{\mathbf{e}}\rangle
&\le \frac{\eta\tilde{K}}{8}\|\nabla F(\mathbf{x})\|^2 + \frac{\eta\tilde{K}L^2}{8}
\|\bar{\mathbf{x}}^{+}-\mathbf{x}\|^2 + \frac{4\,\bar{\varepsilon}^2}{\eta\tilde{K}}.
\end{aligned}
\end{equation*}

\textbf{Step 3 (first-order term).} Recall the trajectory filtration $\mathcal{G}_k$ introduced in the basic notation above, and set $\mathbf{u}_k \coloneqq \frac1N\sum_{i=1}^N \nabla F_i(\mathbf{y}_{i,k})$. We compute $\mathbb{E}[\bar{\mathbf{g}}_k]$ by conditioning twice. (i)~Conditional on $(\mathcal{G}_k, \mathcal{S})$, each $\mathbf{y}_{i,k}$ is $\mathcal{G}_k$-measurable and $\xi_{i,k} \perp \mathcal{G}_k$, so Assumption~\ref{ass:var} gives
\begin{equation*}
\begin{aligned}
\mathbb{E}\bigl[\bar{\mathbf{g}}_k \,\big|\,
\mathcal{G}_k, \mathcal{S}\bigr]
&= \frac1M\sum_{i\in\mathcal{S}}
\mathbb{E}\bigl[\nabla f_i(\mathbf{y}_{i,k};\xi_{i,k})
\,\big|\, \mathcal{G}_k\bigr] \\
&= \frac1M\sum_{i\in\mathcal{S}}
\nabla F_i(\mathbf{y}_{i,k}).
\end{aligned}
\end{equation*}
(ii)~Conditional on $\mathcal{G}_k$ alone, the vectors $\nabla F_i(\mathbf{y}_{i,k})$ are fixed and $\mathcal{S} \perp \mathcal{G}_k$, so the mean part of Lemma~\ref{lem:sampling} applies conditionally and yields $\mathbb{E}[\bar{\mathbf{g}}_k \mid \mathcal{G}_k] = \mathbf{u}_k$. The tower rule then gives $\mathbb{E}[\bar{\mathbf{g}}_k] = \mathbb{E}[\mathbf{u}_k]$. Hence, because $\bar{\mathbf{x}}^{+}-\mathbf{x} = -\eta\sum_k\bar{\mathbf{g}}_k$ and $\nabla F(\mathbf{x})$ is deterministic conditional on $\mathbf{x}$,
\begin{equation*}
\mathbb{E}\langle \nabla F(\mathbf{x}),\, \bar{\mathbf{x}}^{+}-\mathbf{x}\rangle
\;=\; -\eta\sum_{k=0}^{\tilde{K}-1} \mathbb{E}\langle \nabla F(\mathbf{x}),\, \mathbf{u}_k\rangle .
\end{equation*}
(iii)~The polarisation identity $-\langle \mathbf{a},\mathbf{b}\rangle = -\tfrac12\|\mathbf{a}\|^2 - \tfrac12\|\mathbf{b}\|^2 + \tfrac12\|\mathbf{a}-\mathbf{b}\|^2$ with $\mathbf{a} = \nabla F(\mathbf{x})$, $\mathbf{b} = \mathbf{u}_k$ turns each summand into three terms. (iv)~For the last of them, $\nabla F(\mathbf{x}) - \mathbf{u}_k = \frac1N\sum_i\bigl(\nabla F_i(\mathbf{x}) - \nabla F_i(\mathbf{y}_{i,k})\bigr)$, so Lemma~\ref{lem:jensen} and the $L$-smoothness of each $F_i$ (Assumption~\ref{ass:smooth}) give $\|\nabla F(\mathbf{x})-\mathbf{u}_k\|^2 \le \frac{L^2}{N}\sum_i \|\mathbf{x}-\mathbf{y}_{i,k}\|^2$, whose expectation is at most $L^2 d_k$ (Lemma~\ref{lem:drift}). Summing over $k$:
\begin{equation*}
\begin{aligned}
\mathbb{E}\langle \nabla F(\mathbf{x}),
\bar{\mathbf{x}}^{+}-\mathbf{x}\rangle
&\le -\frac{\eta\tilde{K}}{2}\|\nabla F(\mathbf{x})\|^2 - \frac{\eta}{2}\sum_{k}\mathbb{E}\|\mathbf{u}_k\|^2
+ \frac{\eta L^2}{2}\,D.
\end{aligned}
\end{equation*}

\textbf{Step 4 (second moment).} Decompose $\bar{\mathbf{g}}_k = \mathbf{n}_k + \mathbf{v}_k + \mathbf{u}_k$ with sampling noise $\mathbf{n}_k \coloneqq \frac1M\sum_{i\in\mathcal{S}}(\mathbf{g}_{i,k}-\nabla F_i(\mathbf{y}_{i,k}))$ and selection deviation $\mathbf{v}_k \coloneqq \frac1M\sum_{i\in\mathcal{S}}\nabla F_i(\mathbf{y}_{i,k}) - \mathbf{u}_k$; by Lemma~\ref{lem:sumsq} with $n=3$,
$\mathbb{E}\|\bar{\mathbf{x}}^{+}-\mathbf{x}\|^2 \le 3\eta^2\bigl[\,\mathbb{E}\|\sum_k \mathbf{n}_k\|^2 + \mathbb{E}\|\sum_k \mathbf{v}_k\|^2 + \mathbb{E}\|\sum_k \mathbf{u}_k\|^2\bigr]$.

\emph{(a)}~For $k < l$, $\mathbf{n}_k$ is $\sigma(\mathcal{G}_{k+1},\mathcal{S})$-measurable and hence $\sigma(\mathcal{G}_{l},\mathcal{S})$-measurable, while Assumption~\ref{ass:var} gives $\mathbb{E}[\mathbf{n}_l \mid \mathcal{G}_l, \mathcal{S}] = \frac1M\sum_{i\in\mathcal{S}}\mathbb{E}[\mathbf{g}_{i,l} - \nabla F_i(\mathbf{y}_{i,l}) \mid \mathcal{G}_l] = \mathbf{0}$. Therefore, $\mathbb{E}\langle \mathbf{n}_k,\mathbf{n}_l\rangle = \mathbb{E}\bigl\langle \mathbf{n}_k,\, \mathbb{E}[\mathbf{n}_l \mid \mathcal{G}_l,\mathcal{S}]\bigr\rangle = 0$ and $\mathbb{E}\|\sum_k\mathbf{n}_k\|^2 = \sum_k\mathbb{E}\|\mathbf{n}_k\|^2$. For the per-step term, condition on $(\mathcal{G}_k,\mathcal{S})$ and write $\mathbf{n}_k = \frac1M\sum_{i\in\mathcal{S}}\mathbf{z}_i$ with $\mathbf{z}_i \coloneqq \mathbf{g}_{i,k}-\nabla F_i(\mathbf{y}_{i,k})$. The $\mathbf{z}_i$ have conditional mean zero and are mutually independent across clients (the $\xi_{i,k}$ are fresh and independent across $i$, and $\mathcal{S}$ is fixed by the conditioning), so the cross-client terms vanish and $\mathbb{E}[\|\mathbf{n}_k\|^2 \mid \mathcal{G}_k,\mathcal{S}] = \frac{1}{M^2}\sum_{i\in\mathcal{S}}\mathbb{E}[\|\mathbf{z}_i\|^2\mid\mathcal{G}_k] \le \sigma^2/M$ by Assumption~\ref{ass:var}. Sampling without replacement does not affect either computation because $\mathcal{S} \perp \{\xi_{i,k}\}$. Therefore, $\mathbb{E}\|\sum_k \mathbf{n}_k\|^2 \le \tilde{K}\sigma^2/M$.

\emph{(b)}~Conditional on $\mathcal{G}_k$, the vectors $a_i \coloneqq \nabla F_i(\mathbf{y}_{i,k})$ ($i \in [N]$) are fixed with mean $\mathbf{u}_k$, and $\mathcal{S} \perp \mathcal{G}_k$. Applying Lemma~\ref{lem:sampling} conditionally together with $\|a_i - \mathbf{u}_k\| \le 2G$ (Assumption~\ref{ass:clip}) gives $\mathbb{E}[\|\mathbf{v}_k\|^2 \mid \mathcal{G}_k] \le \frac{4G^2}{M}$. Taking expectations and applying Lemma~\ref{lem:sumsq} across the $\tilde{K}$ steps yields $\mathbb{E}\|\sum_k\mathbf{v}_k\|^2 \le \tilde{K}\sum_k \mathbb{E}\|\mathbf{v}_k\|^2 \le \frac{4\tilde{K}^2G^2}{M}$.) 

\emph{(c)}~$\mathbb{E}\|\sum_k\mathbf{u}_k\|^2 \le \tilde{K}\sum_k\mathbb{E}\|\mathbf{u}_k\|^2$. 

\textbf{Step 5 (absorption).} The coefficient multiplying $\mathbb{E}\|\bar{\mathbf{x}}^{+}-\mathbf{x}\|^2$ from Steps 1--2 is $\frac{L}{2} + \frac{\eta\tilde{K}L^2}{8} \le \frac{L}{2}\bigl(1+\frac{1}{32}\bigr) \le \frac{8L}{15}$,

so each Step-4 bracket is multiplied by at most $3\eta^2\cdot\frac{8L}{15} = \frac{8\eta^2 L}{5}$. The mean part is absorbed into Step 3's negative term because $\frac{8\eta^2 L}{5}\tilde{K}\sum_k\mathbb{E}\|\mathbf{u}_k\|^2 \le \frac{\eta}{2}\sum_k\mathbb{E}\|\mathbf{u}_k\|^2$, as $\frac{16}{5}\eta L\tilde{K} \le \frac{16}{5}\cdot\frac18 = \frac{2}{5} \le 1$. We may therefore discard both terms. 

\textbf{Step 6 (collect).} Combining Steps 1--5 and substituting $\frac{\eta L^2}{2}\,D \le \frac{\eta^3 L^2 G^2\tilde{K}^3}{6}$ from Lemma~\ref{lem:drift} yields \eqref{eq:one_round}.

\end{proof}

\subsection{Proofs of the Main Results}\label{app:proof_thm}\label{app:proof_cor}

\begin{proof}[Proof of Theorem~\ref{thm:main}]
Rearranging \eqref{eq:one_round}, summing over $t = 0,\dots,T-1$, telescoping $\mathbb{E} F(\mathbf{w}^{(t)})$, and using $F(\mathbf{w}^{(T)}) \ge F^\ast$ gives $\frac{3\eta\tilde{K}}{8}\sum_t \mathbb{E}\|\nabla F(\mathbf{w}^{(t)})\|^2 \le \Delta + T\,\Xi$, where $\Xi$ is the additive constant in \eqref{eq:one_round}. Dividing by $\frac{3\eta\tilde{K}T}{8}$ multiplies each additive term by $\frac{8}{3\eta\tilde{K}}$ and yields Eq.~\eqref{eq:main_bound}. 

\end{proof}

\begin{proof}[Proof of Corollary~\ref{cor:rate}]
Let $\eta = \min\bigl\{\frac{1}{8L\tilde{K}}, \eta^\star\bigr\}$ with $\eta^\star \coloneqq \sqrt{\frac{M\Delta}{L\tilde{K}T\sigma^2}}$. Using $\frac{1}{\eta} \le 8L\tilde{K} + \frac{1}{\eta^\star}$ and $\eta \le \eta^\star$, the first four terms of Eq.~\eqref{eq:main_bound} are bounded by
\begin{equation*}
\begin{aligned}
&\frac{8\sqrt{L\Delta}\,\sigma}{\sqrt{M\tilde{K}T}}
+ \frac{22\,L\Delta}{T} + 18\,G^2
\sqrt{\frac{L\tilde{K}\Delta}{\sigma^2 M T}} + \frac{M L \tilde{K} G^2 \Delta}{2\,\sigma^2 T}
= \mathcal{O}\Bigl(
\tfrac{\sigma\sqrt{L\Delta}}{\sqrt{M\tilde{K}T}}
\Bigr)
+ \mathcal{O}\Bigl(\tfrac{1}{T}\Bigr).
\end{aligned}
\end{equation*}
For the neighbourhood term, if $\bar{\varepsilon} \le \chi\eta\tilde{K}$, then $\Phi(\bar\varepsilon) \le 11\chi^2 + 2L\eta\tilde{K}\chi^2 \le 11\chi^2 + \frac{\chi^2}{4} \le 12\chi^2$. Under the annealed schedule $\bar{\varepsilon}_t \le \chi_0(t+1)^{-1/4}\eta\tilde{K}$, averaging over $T$ rounds gives $\frac1T\sum_{t=0}^{T-1}\Phi(\bar\varepsilon_t) \le \frac{12\chi_0^2}{T}\sum_{t=1}^{T} t^{-1/2} \le \frac{24\,\chi_0^2}{\sqrt{T}}$.
\end{proof}

\section{Experimental Setup}\label{Apd:ExpSetup}
 
This appendix details the experimental configuration summarised in Section~\ref{sec:experiments}.

\subsection{Datasets}\label{dataset}

We evaluate on two publicly available vision classification benchmarks and one next-character prediction task. FMNIST~\cite{xiao2017fmnist} comprises $60{,}000$ training and $10{,}000$ test greyscale images of $28\times28$ pixels, depicting fashion items across ten classes. CIFAR-10~\cite{Krizhevsky09cifar10} provides $50{,}000$ training and $10{,}000$ test colour images of size $32\times32$, again across ten classes. The Shakespeare dataset~\cite{Caldas2018}, obtained through the Flower Hugging Face repository~\cite{FlowerPaper}, is constructed from \emph{The Complete Works of William Shakespeare} and comprises $4{,}226{,}158$ samples; treating each speaking role in each play as a separate device yields $1{,}129$ clients, which we divide into $904$ training and $225$ evaluation clients. The two vision datasets are therefore heterogeneous by construction and balanced in local sample count, whereas Shakespeare is heterogeneous by nature and severely unbalanced, a distinction that matters for \textsc{FedOAED}, since the trajectory window scales with the number of local batches (Appendix~\ref{app:memory_footprint}).
 
\subsection{Dataset Partitioning}\label{DataPartitioningMethod}
 
The vision datasets are partitioned according to label skew, whereas Shakespeare relies on its inherent user-level structure. We consider five non-IID settings. The hybrid setting, denoted \textsc{IID/Non-IID} in the tables, distributes half of the data identically across clients and skews the remainder. The Dirichlet setting allocates samples according to a Dirichlet distribution whose concentration parameter $\alpha$ governs the degree of skew, with smaller values yielding greater heterogeneity; we fix $\alpha = 0.5$ throughout. The sort-and-partition strategy orders the data by label before dividing it into $C_{\ell}$ chunks, so that $C_{\ell} = 1$ represents the most extreme case in which a client observes a single class; LQ-2 and LQ-3 denote $C_{\ell} = 2, 3$ respectively. Finally, the natural identifier partitioner preserves the user-based segmentation already present in the raw data and is applied exclusively to Shakespeare, where it captures realistic statistical heterogeneity. Further detail on these strategies is available in~\cite{LiPartitioning} and the Flower documentation~\cite{FlowerPaper}.
 
\subsection{Models and Federated Configuration}
 
LeNet-5~\cite{Lecun1998} is adopted for FMNIST ($d \approx 6.2\times10^{4}$ parameters), and ResNet-18~\cite{resnet} for CIFAR-10 ($d \approx 1.1\times10^{7}$), with batch normalisation replaced by group normalisation~\cite{Hsieh2019}. The substitution is standard practice in federated learning, where small and heterogeneous local batches make batch statistics unreliable; it also has a secondary benefit here, in that removing the running statistics leaves the trainable parameters as the entire model state, so the flattened snapshot of Section~\ref{sec:ProposedAlgorithm} captures everything. Shakespeare is modelled with a GRU-based recurrent network~\cite{Chung2014} consisting of an eight-dimensional character embedding, a single recurrent layer with $128$ hidden units, and a fully connected layer mapping the final hidden state onto the character vocabulary ($d \approx 6.7\times10^{4}$).

\subsection{Client Configuration}
FMNIST is distributed across $500$ clients, CIFAR-10 across $250$, and Shakespeare across $904$ training clients. Each federated round samples five clients uniformly at random without replacement, corresponding to participation rates of $1\%$, $2\%$ and below $1\%$ respectively. Local sample counts follow directly: approximately $120$ samples per client on FMNIST and $200$ on CIFAR-10, against a mean of roughly $3{,}700$ on Shakespeare with substantial variance across roles. The test data follow the training partitions: the vision models are evaluated on the full client pool after each round, while Shakespeare is evaluated on a held-out set of the remaining $225$ clients. Round counts vary by dataset and partition and are given in Table~\ref{tab:rounds}; all reported accuracies and losses are averaged over the final $10\%$ of rounds.

\begin{table*}[t]
\centering
\small{
\begin{tabular}{lccc}
	\toprule
	& \multicolumn{3}{c}{\textbf{Dataset}} \\
	\cmidrule(lr){2-4}
	\textbf{Parameter} & \textbf{F-MNIST} & \textbf{CIFAR-10} & \textbf{Shakespeare} \\
	\midrule
	\# of clients & 500 & 250 & 904 \\
	\# of participating clients & 5 & 5 & 5 \\
	\% of participating clients & 1\% & 2\% & $\leq$ 1\% \\
	\# of clients for evaluation & 500 & 250 & 225 \\
	Batch size $B$ & 20 & 20 & 64 \\
	Local epochs $E$ & 3 & 5 & 5 \\
	Avg. Batches per client $b_i$ & $\approx$6 & $\approx$10 & $\approx$58 \\
	Client learning rate \(\eta_c\) & 0.01 & 0.01 & $\{0.1, 0.01\}$ \\
	Client momentum & 0.9 & 0.9 & 0.9 \\
	Local model & LeNet-5 & ResNet-18 & GRU-RNN \\
	Model dimension $d$ & $6.2\!\times\!10^{4}$ & $1.1\!\times\!10^{7}$ & $6.7\!\times\!10^{4}$ \\
	\# of seeds & 3 & 1 & 1 \\
	\bottomrule
\end{tabular}
}
\caption{Experimental setup. $b_i$ is derived from the local sample count and batch size and is reported because it determines the trajectory window $W = \lfloor \gamma b_i \rfloor$.}
\label{tab:ExperimentalDetails}
\end{table*}

\begin{table*}[t]
    \centering
    \small{
    \begin{tabular}{lccccc}
        \toprule
        & \multicolumn{5}{c}{\textbf{Partition}} \\
        \cmidrule(lr){2-6}
        \textbf{Dataset} & \textbf{IID/Non-IID} & \textbf{Dirichlet} & \textbf{LQ-2} & \textbf{LQ-3} & \textbf{Natural} \\
        \midrule
        F-MNIST     & 100 & 150 & 150 & 150 & --- \\
        CIFAR-10    & 200 & 300 & 400 & 300 & --- \\
        Shakespeare & --- & --- & --- & --- & 200 \\
        \bottomrule
    \end{tabular}
    }
    \caption{Number of training rounds by dataset and partition strategy. A dash
    indicates a partition not applicable to that dataset. All reported accuracies
    and losses are averaged over the final $10\%$ of rounds.}
    \label{tab:rounds}
\end{table*}
 
\begin{table*}[t]
\centering
\small{
\begin{tabular}{lccc}
    \toprule
    & \multicolumn{3}{c}{\textbf{Dataset}} \\
    \cmidrule(lr){2-4}
    \textbf{Parameter} & \textbf{F-MNIST} & \textbf{CIFAR-10} & \textbf{Shakespeare} \\
    \midrule
    AE architecture & \multicolumn{3}{c}{(Progressive, Residual)} \\
    AE latent width & 16 & (16, 32) & 32 \\
    AE hidden width $h$ & 128 & (128, 256) & 256\\
    Chunk fraction $\phi$ & \multicolumn{3}{c}{0.01} \\
    Chunk width $C = \lfloor \phi d \rfloor$ & 617 & 111{,}739 & 666 \\
    Refine cadence $E_r$ & \multicolumn{3}{c}{1} \\
    Mixing coefficient $\lambda$ & \multicolumn{3}{c}{0.01} \\
    AE epochs $e$ & \multicolumn{3}{c}{10} \\
    Snapshot interval $s$ & \multicolumn{3}{c}{1} \\
    Buffer capacity $Q$ & \multicolumn{3}{c}{100} \\
    Window percentage $\gamma$ & \multicolumn{3}{c}{0.8} \\
    Realised window $W_i$ & $\approx$4 & $\approx$8 & $\approx$46 \\
    Rel.\ error threshold $\tau$ & \multicolumn{3}{c}{0.3} \\
    Min.\ snapshots $m$ & \multicolumn{3}{c}{3} \\
    Use loss-weighted AE & \multicolumn{3}{c}{True} \\
    Loss temperature $\kappa$ & \multicolumn{3}{c}{1.0} \\
    Loss tolerance $\rho$ & \multicolumn{3}{c}{0.1} \\
    Gating batches $b$ & \multicolumn{3}{c}{5} \\
    Recovery steps $R$ & \multicolumn{3}{c}{3} \\
    Max gradient norm $G_{\mathrm{clip}}$ & \multicolumn{3}{c}{0.5} \\
    Stability constant $\epsilon_0$ & \multicolumn{3}{c}{$10^{-8}$} \\
    AE learning rate $\eta_a$ & \multicolumn{3}{c}{0.001} \\
    Chunk mini-batch & \multicolumn{3}{c}{512} \\
    \bottomrule
\end{tabular}
}
\caption{\textsc{FedOAED} hyperparameter setup. Parenthesised pairs denote a grid searched over both values.}
\label{tab:FedOAEDExperimentalDetails}
\end{table*}

\section{Implementation Details and Reproducibility}
\label{app:implementation}

\subsection{Autoencoder Architectures}
\label{app:ae_arch}
 
Both architectures map $\mathbb{R}^{C} \to \mathbb{R}^{C}$, where $C = \lfloor \phi d \rfloor$ is the chunk width, and are applied to all $P = \lceil d/C \rceil$ chunks with shared weights in a single batched forward pass. The argument \texttt{param\_dim} in the listings therefore denotes the chunk width $C$, not the model dimension $d$.

In all FMNIST experiments, we instantiate these autoencoders with
$\text{hidden}=128$ and $\text{latent\_dim}=16$. For the CIFAR-10 dataset, we use
a parameter grid of $\text{hidden} \in \{128, 256\}$ and
$\text{latent\_dim} \in \{16, 32\}$
(Table~\ref{tab:FedOAEDExperimentalDetails}). In addition,
$\text{hidden}=256$ and $\text{latent\_dim}=32$ were set for the Shakespeare
dataset.

\subsection{Numerical Safeguards}
\label{app:guards}

The refinement is skipped for the current cycle, leaving $\mathbf{w}_i$ untouched, under any of the following conditions: $\theta_i$ is non-finite; the window statistics $(\mathbf{m}_i, \mathbf{s}_i)$ are unavailable; the reconstruction $\widehat{\theta}_i$ or the mixed vector $\widetilde{\theta}_i$ is non-finite after clamping; or the effective mixing coefficient $c\lambda$ evaluates to zero. Reconstructions have \texttt{NaN} entries replaced by zero and are clamped element-wise to $\pm 10^{6}$ before mixing. Non-finite outputs or losses encountered during local SGD abort the round for that client rather than being silently absorbed. Each path is logged separately, so a rejection attributable to the relative-norm gate can be distinguished from a numerical failure.
 
\subsection{Autoencoder Training and State}
\label{app:ae_state}

The autoencoder is trained using Adam with a learning rate of $10^{-3}$ and no weight decay over shuffled mini-batches of $512$ chunks, minimising the $\omega$-weighted MSE renormalised within each mini-batch. Within a round, the autoencoder weights persist across refinement cycles while a fresh Adam optimiser is instantiated at each cycle. The window statistics $(\mathbf{m}_i, \mathbf{s}_i)$ and the gating set $\widehat{\mathcal{D}}_i$ are likewise recomputed at every cycle. Because the initial epoch budget $e$ applies only to the first cycle of a round, total autoencoder training per round is $e + (\lfloor E/E_r \rfloor - 1)$ epochs.
 
\subsection{Recovery Steps and Short-Shard Handling}
\label{app:divergences}

In our implementation, the $R$ post-refinement recovery steps execute only when the rollback gate accepts, whereas the theoretical model in Section~\ref{sec:theory} executes them unconditionally to keep the step count $\tilde{K} = K + JR$ deterministic across clients.

Additionally, for datasets with heavily skewed local sample counts (such as the shortest speaking roles in Shakespeare, where $b_i \le \lceil m/\gamma \rceil$), $\lfloor \gamma b_i \rfloor$ can fall below the minimum snapshot count $m$. Whenever a client's effective window contains fewer than $m$ snapshots, refinement is skipped (or rejected by the relative-norm gate) and the client defaults to a standard FedAvg update for that round. For general deployments with heavy-tailed client shard sizes, clamping the window via $W \gets \min\!\bigl(Q, \max(m, \lfloor \gamma b_i \rfloor)\bigr)$ ensures at least $m$ snapshots are retained whenever $|\mathcal{B}| \ge m$.

\begin{algorithm}[t]
\caption{Residual}
\label{alg:modern-autoencoder}
\small{
\begin{algorithmic}[1]
\Require $\text{param\_dim}$ (chunk width $C$)
\Require $\text{latent\_dim}$ (default: $32$)
\Require $\text{hidden}$ (default: $256$)
\Require $\text{num\_blocks}$ (default: $2$)
\Require $\text{expansion\_factor}$ (default: $2$)
 
\Function{ResBlock}{$u$}
    \State $v \leftarrow \text{LayerNorm}(u)$
    \State $v \leftarrow \text{Linear}(\text{hidden} \rightarrow \text{hidden} \times \text{expansion\_factor})(v)$
    \State $v \leftarrow \text{GELU}(v)$
    \State $v \leftarrow \text{Linear}(\text{hidden} \times \text{expansion\_factor} \rightarrow \text{hidden})(v)$
    \State \Return $u + v$
\EndFunction
 
\State \textbf{Encoder } $E$: Linear($\text{param\_dim} \rightarrow \text{hidden}$)
\State \quad\quad then sequentially apply \textsc{ResBlock} $\text{num\_blocks}$ times
\State \quad\quad then LayerNorm($\text{hidden}$)
\State \quad\quad then Linear($\text{hidden} \rightarrow \text{latent\_dim}$)
 
\State \textbf{Decoder } $D$: Linear($\text{latent\_dim} \rightarrow \text{hidden}$)
\State \quad\quad then sequentially apply \textsc{ResBlock} $\text{num\_blocks}$ times
\State \quad\quad then LayerNorm($\text{hidden}$)
\State \quad\quad then Linear($\text{hidden} \rightarrow \text{param\_dim}$)
 
\Function{Forward}{$x$}
    \If{$\text{dim}(x) = 1$}
        \State $x \leftarrow \text{unsqueeze}(x, \;0)$
        \State $\text{was\_vector} \leftarrow \text{true}$
    \Else
        \State $\text{was\_vector} \leftarrow \text{false}$
    \EndIf
 
    \State $z \leftarrow E(x)$
    \State $\hat{x} \leftarrow D(z)$
 
    \If{$\text{was\_vector}$}
        \State \Return $\text{squeeze}(\hat{x},\;0)$
    \Else
        \State \Return $\hat{x}$
    \EndIf
\EndFunction
\end{algorithmic}
}
\end{algorithm}
 
\begin{algorithm}[t]
\caption{Progressive Bottleneck Autoencoder}
\label{alg:progressive-autoencoder}
\small{
\begin{algorithmic}[1]
\Require $\text{param\_dim}$ (chunk width $C$)
\Require $\text{latent\_dim}$ (default: $32$)
\Require $\text{hidden}$ (default: $256$)
 
\State $\text{inter\_dim} \leftarrow \max(\text{hidden}, \min(\lfloor\text{param\_dim} / 2\rfloor, \text{hidden} \times 4))$
 
\State \textbf{Encoder } $E$:
\State \quad\quad Linear($\text{param\_dim} \rightarrow \text{inter\_dim}$), then BatchNorm1d, then GELU
\State \quad\quad then Linear($\text{inter\_dim} \rightarrow \text{hidden}$), then BatchNorm1d, then GELU
\State \quad\quad then Linear($\text{hidden} \rightarrow \text{latent\_dim}$)
 
\State \textbf{Decoder } $D$:
\State \quad\quad Linear($\text{latent\_dim} \rightarrow \text{hidden}$), then BatchNorm1d, then GELU
\State \quad\quad then Linear($\text{hidden} \rightarrow \text{inter\_dim}$), then BatchNorm1d, then GELU
\State \quad\quad then Linear($\text{inter\_dim} \rightarrow \text{param\_dim}$)
 
\Function{Forward}{$x$}
    \If{$\text{dim}(x) = 1$}
        \State $x \leftarrow \text{unsqueeze}(x, \;0)$
        \State $\text{was\_vector} \leftarrow \text{true}$
    \Else
        \State $\text{was\_vector} \leftarrow \text{false}$
    \EndIf
 
    \State $z \leftarrow E(x)$
    \State $\hat{x} \leftarrow D(z)$
 
    \If{$\text{was\_vector}$}
        \State \Return $\text{squeeze}(\hat{x},\;0)$
    \Else
        \State \Return $\hat{x}$
    \EndIf
\EndFunction
\end{algorithmic}
}
\end{algorithm}
 
\section{Full Experimental Results}\label{Apd:ExpResults}
 
Figure~\ref{fig:final_evaluation_accuracy} in the main text reports accuracy trajectories; Figure~\ref{fig:final_evaluation_loss} below gives the corresponding generalisation losses, and Table~\ref{tab:partition_results} the final values for the vision benchmarks.

Across the vision benchmarks the ordering is consistent: \textsc{FedOAED} attains the best final accuracy in seven of the eight dataset--partition combinations, with the sole exception of CIFAR-10 LQ-3, where FedProx is $0.80$ percentage points ahead. The margin over FedAvg widens as heterogeneity increases on FMNIST ($2.83$ points on IID/Non-IID, rising to $3.78$ on LQ-3), which is the behaviour the method is designed to produce. The two $\mathcal{O}(Nd)$ caching baselines behave very differently from one another: FedVARP is competitive throughout, whereas MIFA is unstable under our participation rates, falling to $28.56\%$ on FMNIST Dirichlet and $23.78\%$ on CIFAR-10 IID/Non-IID. Its large standard deviations on FMNIST indicate that the instability is not a single unlucky run. SCAFFOLD, whose client control variates are updated only about once every hundred rounds at $1\%$ participation, is consistently the weakest drift-correction baseline, illustrating the staleness issue noted in Section~\ref{sec:intro}.

Shakespeare is reported separately in Table~\ref{tab:shakespeare_results}, as it has a single natural partition rather than four synthetic splits and the client learning rate was swept over $\{0.1, 0.01\}$ for every method. The ordering is stable across both rates: FedVARP and \textsc{FedOAED} occupy the top two positions, separated by $0.07$ points at $\eta_c = 0.1$ and $0.51$ at $\eta_c = 0.01$, with a clear margin to third place ($1.92$ points at the lower rate). \textsc{FedOAED} is therefore the only method to match FedVARP on this benchmark, and does so whilst leaving server storage at $\mathcal{O}(d)$ against FedVARP's $\mathcal{O}(Nd)$.

The failure modes of SCAFFOLD ($20.12\%$ and $15.85\%$) and MIFA (below $15\%$ at both learning rates) directly reflect the low-participation regime. At five clients per round, Shakespeare accumulates only $1{,}000$ client visits across $904$ clients over $200$ rounds (and F-MNIST accumulates $750$ visits across $500$ clients over $150$ rounds), meaning a typical client is sampledRoughly once during training. Consequently, persisted client control variates or cached server pseudo-gradients become severely outdated between visits, whereas \textsc{FedOAED} remains unaffected because it is stateless across rounds.

The small remaining gap to FedVARP on Shakespeare may also reflect the fixed window fraction $\gamma = 0.8$ used across all benchmarks. Because Shakespeare clients average $b_i \approx 58$ batches per epoch, $\gamma = 0.8$ yields a window of $W_i \approx 46$ snapshots (compared with $4$ on F-MNIST and $8$ on CIFAR-10). As shown in Appendix~\ref{app:sensitivity_analysis}, shorter windows ($\gamma = 0.5$) improve accuracy by avoiding earlier, higher-loss trajectory states, suggesting that a smaller $\gamma$ on long-epoch tasks could further benefit \textsc{FedOAED}.

\begin{figure*}[t]
	\centering
	\includegraphics[scale=0.105]{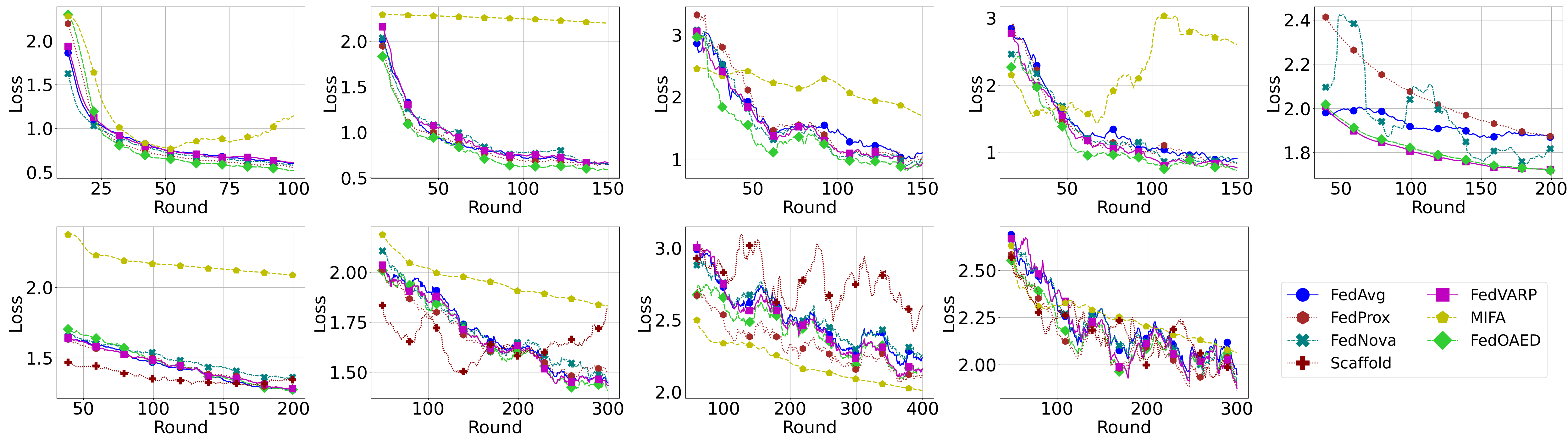}
	\caption{Loss comparison across various data partitioning methods. The top row (left to right) illustrates IID/Non-IID, Dirichlet, LQ-2, and LQ-3 splits on the FMNIST dataset, followed by the IdPartitioner on the Shakespeare dataset. The bottom row depicts the corresponding IID/Non-IID, Dirichlet, LQ-2, and LQ-3 configurations on the CIFAR-10 dataset. SCAFFOLD and MIFA are omitted from the Shakespeare panel, where their losses fluctuate over a range that compresses the remaining curves.}
	\label{fig:final_evaluation_loss}
\end{figure*}
 
\begin{table*}[httb]
    \centering
    \small{
    \begin{tabular}{@{}llcccc@{}}
    \toprule
    \textbf{Strategy} & \textbf{Dataset} & \textbf{IID-NonIID} & \textbf{Dirichlet} & \textbf{LQ-2} & \textbf{LQ-3} \\ \midrule
    FedAvg            & FMNIST           & $76.73 \pm 0.34$    & $75.28 \pm 0.81$   & $66.28 \pm 0.40$ & $68.97 \pm 2.07$ \\
    FedProx           & FMNIST           & $77.95 \pm 1.23$    & $75.36 \pm 0.21$   & $64.69 \pm 0.45$ & $69.62 \pm 0.57$ \\
    SCAFFOLD          & FMNIST           & $63.96 \pm 2.05$    & $62.48 \pm 3.67$   & $45.61 \pm 4.12$ & $54.51 \pm 2.38$ \\
    FedNova           & FMNIST           & $77.07 \pm 0.94$    & $75.70 \pm 0.76$   & $66.45 \pm 1.82$ & $67.80 \pm 0.58$ \\
    FedVARP           & FMNIST           & $76.08 \pm 0.19$    & $75.44 \pm 0.89$   & $65.68 \pm 1.40$ & $70.30 \pm 2.60$ \\
    MIFA              & FMNIST           & $64.77 \pm 2.64$    & $28.56 \pm 11.22$  & $29.56 \pm 7.70$ & $35.23 \pm 4.82$ \\
    \textsc{FedOAED}  & FMNIST           & $\mathbf{79.56 \pm 0.97}$ & $\mathbf{78.12 \pm 0.72}$ & $\mathbf{69.53 \pm 2.35}$ & $\mathbf{72.75 \pm 1.18}$ \\
    \midrule
    FedAvg            & CIFAR-10         & $62.05$ & $57.51$ & $34.54$ & $42.35$ \\
    FedProx           & CIFAR-10         & $62.15$ & $56.60$ & $36.62$ & $\mathbf{45.59}$ \\
    SCAFFOLD          & CIFAR-10         & $60.22$ & $50.31$ & $32.35$ & $39.03$ \\
    FedNova           & CIFAR-10         & $59.78$ & $57.30$ & $33.38$ & $43.35$ \\
    FedVARP           & CIFAR-10         & $62.30$ & $57.69$ & $36.42$ & $43.95$ \\
    MIFA              & CIFAR-10         & $23.78$ & $31.95$ & $27.20$ & $24.72$ \\
    \textsc{FedOAED}  & CIFAR-10         & $\mathbf{62.85}$ & $\mathbf{58.39}$ & $\mathbf{37.29}$ & $44.79$ \\
    \bottomrule
    \end{tabular}
    }
    \caption{Performance comparison of various strategies across the vision dataset partitions. Top-1 accuracy (\%) averaged over the final $10\%$ of rounds (Table~\ref{tab:rounds}); mean $\pm$ standard deviation where multiple seeds were run. Best result per column in bold. Shakespeare, which has a single natural partition, is reported separately in Table~\ref{tab:shakespeare_results}.}
    \label{tab:partition_results}
\end{table*}

\begin{table*}[hbt!]
    \centering
    \small{
    \begin{tabular}{lcc}
        \toprule
        \textbf{Method} & $\boldsymbol{\eta_c = 0.1}$ & $\boldsymbol{\eta_c = 0.01}$ \\
        \midrule
        FedAvg           & 47.98 & 47.29 \\
        FedProx          & 45.88 & 45.04 \\
        SCAFFOLD         & 20.12 & 15.85 \\
        FedNova          & 41.69 & 47.88 \\
        MIFA             & 13.72 & 14.29 \\
        FedVARP          & 48.43 & \textbf{50.31} \\
        \textsc{FedOAED} & 48.36 & 49.80 \\
        \bottomrule
    \end{tabular}
    }
    \caption{Next-character prediction accuracy (\%) on the Shakespeare dataset (Natural IdPartitioner, $904$ training and $225$ evaluation clients) at both client learning rates, averaged over the final $10\%$ of rounds. Single seed.}
    \label{tab:shakespeare_results}
\end{table*}

\section{Computational and Memory Profile}
\label{app:compute_profile}
 
A core design objective of \textsc{FedOAED} is to eliminate the prohibitive $\mathcal{O}(Nd)$ server-side memory overhead required by variance-reduction baselines such as MIFA and FedVARP. By shifting the denoising burden to the client device, \textsc{FedOAED} achieves a strict $\mathcal{O}(d)$ server footprint. This design trades local computation and transient client memory for server scalability; below we quantify these on-device costs and identify when they become limiting.
 
\subsection{Local Computational Overhead}

Empirically, training the local autoencoder on the trajectory window introduces a computational overhead. In our PyTorch implementation, profiling on a single device reveals that one full local update cycle of \textsc{FedOAED} takes approximately $3\times$ the wall-clock time of a standard FedAvg update. This overhead stems directly from the forward and backward passes required to incrementally train the autoencoder ($\mathcal{A}_i$) over the sliding window of $W_i$ snapshots.

In cross-device federated learning, however, local computation is rarely the primary bottleneck; communication latency and server scalability dominate. Two considerations bear on this penalty. First, network transmission over fluctuating wireless connections is significantly slower and more power-intensive than local FLOPs, so wall-clock cost per round is dominated by communication rather than by the autoencoder; any reduction in the rounds needed to reach a target accuracy would therefore offset the local penalty, though we fix the round budget across methods and do not measure this. Second, the overhead is highly tunable. Autoencoder training cost is invariant in the chunk percentage $\phi$ (the chunk count $1/\phi$ and chunk width $\phi d$ cancel), so compute is scaled instead through the refinement cadence $E_r$, the initial autoencoder epochs $e$, and the window percentage $\gamma$; $\phi$ independently controls the autoencoder's size, and $s$ and $Q$ the trajectory buffer. Any of these can be relaxed for resource-constrained edge devices without altering the server aggregation logic.
 
\subsection{Autoencoder Size and Its Scaling in $d$}
\label{app:ae_scaling}

Chunking is essential to make the on-device denoiser tractable. A full-width autoencoder ingesting $\mathbb{R}^d$ directly would require $\mathcal{O}(dh)$ projection parameters, roughly $5.6\times10^{9}$ parameters for ResNet-18 at $h = 256$, which cannot be fit on-device or trained from a few dozen trajectory snapshots. Sharing a single autoencoder across chunks of width $C = \lfloor \phi d \rfloor$ reduces the projection layers to $\mathcal{O}(\phi d h)$ (a $100\times$ reduction at $\phi = 0.01$) while expanding the effective training set from $W_i$ snapshots to $W_i / \phi$ chunk examples.

Because $\phi$ is defined as a fixed fraction of $d$, the chunk width $C$ still scales linearly with the backbone dimension $d$ (Table~\ref{tab:ae_size}). The parameter count is dominated by the outer $C \times h$ projections, whereas the latent dimension $z$ changes the total by less than $0.1\%$. On small backbones (F-MNIST and Shakespeare), fixed intermediate layers dominate the relative ratio, whereas on ResNet-18 the absolute parameter count ranges from $2.9\times10^{7}$ (Residual, $h=128$) to $2.3\times10^{8}$ (Progressive, $h=256$). For larger backbones, $\phi$ can be scaled down inversely with $d$, equivalent to fixing an absolute chunk width such as $C = 1024$, so that the autoencoder's parameter count remains constant regardless of backbone size.

\begin{table*}[hbt!]
    \centering
    \small
    \begin{tabular}{lcccc}
        \toprule
        \textbf{Dataset} & \textbf{$C$} & \textbf{$h$} &
        \textbf{Residual} & \textbf{Progressive} \\
        \midrule
        F-MNIST     & 617       & 128 & $4.3\!\times\!10^{5}$ (6.9$\times$) & $4.7\!\times\!10^{5}$ (7.6$\times$) \\
        \midrule
        \multirow{2}{*}{CIFAR-10}    & \multirow{2}{*}{111{,}739} & 128 & $2.9\!\times\!10^{7}$ (2.6$\times$) & $1.1\!\times\!10^{8}$ (10.3$\times$) \\
                    &           & 256 & $5.8\!\times\!10^{7}$ (5.2$\times$) & $2.3\!\times\!10^{8}$ (20.5$\times$) \\
        \midrule
        Shakespeare & 666       & 256 & $1.4\!\times\!10^{6}$ (21.2$\times$) & $6.3\!\times\!10^{5}$ (9.5$\times$) \\
        \bottomrule
    \end{tabular}
    \caption{Autoencoder size relative to the backbone, for both architectures. Chunk width $C = \lfloor \phi d \rfloor$ with $\phi = 0.01$. Counts are governed by the hidden width $h$; the latent width $z$ changes them by under $0.1\%$ and is omitted.}
    \label{tab:ae_size}
\end{table*}

\begin{table*}[hbt!]
    \centering
    \small
    \begin{tabular}{lcccc}
        \toprule
        \textbf{Dataset} & \textbf{$E b_i / s$} & \textbf{Snapshots held} & \textbf{$Q$ binds?} & \textbf{Size} \\
        \midrule
        F-MNIST     & 18  & 18  & no  & $\approx\!4$\,MB \\
        CIFAR-10    & 50  & 50  & no  & $\approx\!2.1$\,GB \\
        Shakespeare & 290 & 100 & yes & $\approx\!25$\,MB \\
        \bottomrule
    \end{tabular}
    \caption{Realised trajectory-buffer occupancy at float32 precision.
    $Q = 100$ and $s = 1$ throughout.}
    \label{tab:memory}
\end{table*}

\begin{table*}[hbt!]
    \centering
    \small{
    \begin{tabular}{lcc}
        \toprule
        \textbf{Method} & \textbf{Avg. Accuracy} & \textbf{Avg. Loss} \\
        \midrule
        FedAvg (Baseline) & $75.28 \pm 0.81$ & $0.655 \pm 0.047$ \\
        \textsc{FedOAED} (No RG) & $77.04 \pm 0.83$ & $0.641 \pm 0.037$ \\
        \textsc{FedOAED} (No LWS) & $78.10 \pm 0.79$ & $0.605 \pm 0.027$ \\
        \textbf{\textsc{FedOAED} (Full)} & $\mathbf{78.65} \pm \mathbf{0.72}$ & $\mathbf{0.585} \pm \mathbf{0.019}$ \\
        \bottomrule
    \end{tabular}
    }
    \caption{Ablation study on FMNIST (Dirichlet $\alpha=0.5$) over three random seeds, isolating the impact of Loss-Weighted Sampling (LWS) and the Rollback Gate (RG). Mean $\pm$ standard deviation across seeds of the test accuracy and generalisation loss, each averaged over the final $10\%$ of rounds (Table~\ref{tab:rounds}). Removing either component degrades both metrics, and the full method is the strongest configuration on both. The \textsc{FedOAED} rows were run as a separate batch from the comparison of Table~\ref{tab:partition_results}, whilst the FedAvg baseline is carried over from it; the \textsc{FedOAED} (Full) configuration is identical in both, and the difference in reported means ($78.65$ here against $78.12$ there) reflects run-to-run variation and lies well within the seed standard deviation of either.}
    \label{tab:ablation_safety}
\end{table*}

\subsection{Client-Side Memory Footprint}
\label{app:memory_footprint}

On-device memory has two components. The autoencoder and its optimiser state are addressed above. The trajectory buffer holds $\min\!\left(Q,\, E b_i / s\right)$ flattened parameter vectors and therefore occupies $\mathcal{O}\!\left(\min(Q, E b_i/s)\, d\right)$. Both components are linear in $d$; which dominates depends on the configuration, and on the largest backbone we study the buffer does. Table~\ref{tab:memory} gives the realised figures.
 
The three datasets illustrate three distinct regimes. On FMNIST the buffer is small in absolute terms, though still roughly twice the size of the autoencoder. On Shakespeare the buffer capacity $Q$ binds: with $b_i \approx 58$ and $E = 5$ the client would otherwise accumulate close to three hundred snapshots, so first-in-first-out eviction is active throughout the round and $Q$ is performing exactly the function it was introduced for. On CIFAR-10 neither $Q$ nor eviction binds, but each snapshot is over $180$ times larger, and the buffer reaches a size that is unproblematic in the simulation under which our results were obtained yet exceeds the memory available on a typical mobile device.
 
Three mitigations reduce the buffer term without touching the server or the aggregation logic: raising the snapshot interval $s$, which subsamples the trajectory and divides the term directly; lowering $Q$, which caps the buffer outright and, since our sensitivity analysis indicates shorter windows cost no accuracy, can be reduced towards the window length actually used; and storing snapshots at reduced precision, since they are consumed only to compute an element-wise mean and deviation. We have not measured the accuracy cost of the third, and note it in Appendix~\ref{app:limitations}.
 
In all cases the buffer and $\mathcal{A}_i$ are transient: they are allocated by the $|\mathcal{S}^{(t)}|$ sampled clients at the start of a round and released at its end. This contrasts with the $\mathcal{O}(Nd)$ server-side state of the caching methods (e.g. FedVARP, MIFA), which persists for the duration of training and scales with the full client population rather than the sampled subset.

\section{Ablation Studies: Safety and Trajectory Mechanisms}
\label{app:ablation}
 
To understand the individual contributions of \textsc{FedOAED}'s core mechanisms, we conducted an ablation study isolating Loss-Weighted Sampling (LWS) and the Rollback Gate (RG). Table~\ref{tab:ablation_safety} presents the results under the highly heterogeneous Dirichlet ($\alpha = 0.5$) partition on the FMNIST dataset, averaged over three seeds. Both mechanisms contribute in the same direction, and the ordering is consistent across accuracy and loss: the rollback gate is by some margin the more important of the two, and removing either leaves a configuration that still improves on FedAvg.

\subsection{Impact of the Rollback Gate}

Disabling the rollback gate degrades performance on both metrics. Autoencoders operating on high-dimensional parameter chunks can occasionally produce reconstructions that, while structurally close in Euclidean space (passing the relative-norm gate $\tau$), land in sharp, high-loss regions of the local optimisation landscape. Without the deterministic pre- and post-refinement loss evaluation on identical cached batches ($\widehat{\mathcal{D}}_i$), the client blindly commits to these destructive updates. As shown in Table~\ref{tab:ablation_safety}, removing the rollback gate costs $1.61$ accuracy points ($78.65 \to 77.04\%$) and raises the generalisation loss from $0.585$ to $0.641$, making it the more consequential of the two mechanisms by a wide margin. Both gaps are roughly twice the seed variance of either configuration, and the accuracy and loss effects agree in direction and rough magnitude, though at three seeds neither reaches conventional significance.
 
\subsection{Impact of Loss-Weighted Sampling}

Without the temperature-scaled softmax weighting, the autoencoder minimises the reconstruction error uniformly across the sliding window. An unweighted autoencoder therefore centres its reconstruction on the mean of the window rather than on its lowest-loss members, which is the wrong target when those members differ in quality, whether because the window spans a descending segment of the trajectory or because mini-batch noise makes some snapshots better-placed than others. By re-weighting the objective with $\omega_j \propto \exp(-z_j/\kappa)$, LWS steers the autoencoder toward lower-loss iterates rather than weighting all trajectory states equally. Removing this mechanism costs $0.55$ accuracy points ($78.65 \to 78.10\%$) and raises the loss from $0.585$ to $0.605$, a smaller effect than the rollback gate, and one that does not clearly exceed seed variance at three seeds, though it is consistent in direction across both metrics.

\subsection{What the Ablations Do Not Isolate}
\label{app:ablation_caveats}

Two components are not separated by this study. The post-refinement recovery steps are not ablated independently: with $E_r = 1$ and $R = 3$ a client performs $JR = 9$ recovery steps against $K = 18$ regular steps on FMNIST, a fifty per cent increase in local optimisation, so part of the gain over FedAvg could in principle be attributed to additional local optimisation rather than to denoising as such. The mixing-coefficient sweep in Table~\ref{tab:sensitivity_analysis} provides indirect evidence here, since $\lambda$ varies while the recovery budget $R$ stays fixed (Appendix~\ref{app:sweep_reading}), although the actual number of recovery steps still depends on the gate's acceptance rate. Directly isolating the denoising gain would require comparing against FedAvg with $\tilde{K} = K + JR$ local steps, which we leave for future work. Nor is chunk-level weight sharing ablated, since a full-width autoencoder is infeasible at the model sizes considered here; that comparison would require a different experimental design.

\begin{table*}[t]
    \centering
    
    \small{
    \begin{tabular}{llcc}
        \toprule
        \textbf{Parameter Swept} & \textbf{Tested Value} & \textbf{Avg. Accuracy} & \textbf{Avg. Loss} \\
        \midrule
        \multirow{5}{*}{\shortstack[l]{Mixing Coefficient\\($\lambda$)}} 
        & 0.10 & 79.39 $\pm$ 0.63\% & 0.571 $\pm$ 0.021 \\
        & 0.05 & 79.45 $\pm$ 0.51\% & 0.567 $\pm$ 0.005 \\
        & 0.01 (Base) & 78.65 $\pm$ 0.72\% & 0.585 $\pm$ 0.019 \\
        & 0.005 & 78.50 $\pm$ 0.83\% & 0.595 $\pm$ 0.036 \\
        & 0.001 & 77.84 $\pm$ 0.90\% & 0.602 $\pm$ 0.032 \\
        \midrule
        \multirow{3}{*}{\shortstack[l]{Relative Threshold\\($\tau$)}} 
        & 0.1 & 77.85 $\pm$ 0.62\% & 0.592 $\pm$ 0.015 \\
        & 0.3 (Base) & 78.65 $\pm$ 0.72\% & 0.585 $\pm$ 0.019 \\
        & 0.5 & 78.93 $\pm$ 0.48\% & 0.571 $\pm$ 0.011 \\
        \midrule
        \multirow{3}{*}{\shortstack[l]{Loss Tolerance\\($\rho$)}} 
        & 0.05 & 78.52 $\pm$ 0.55\% & 0.582 $\pm$ 0.014 \\
        & 0.1 (Base) & 78.65 $\pm$ 0.72\% & 0.585 $\pm$ 0.019 \\
        & 0.2 & 78.11 $\pm$ 0.68\% & 0.590 $\pm$ 0.018 \\
        \midrule
        \multirow{3}{*}{\shortstack[l]{Loss Temperature\\($\kappa$)}} 
        & 0.5 & 77.94 $\pm$ 0.59\% & 0.593 $\pm$ 0.016 \\
        & 1.0 (Base) & 78.65 $\pm$ 0.72\% & 0.585 $\pm$ 0.019 \\
        & 2.0 & 78.43 $\pm$ 0.47\% & 0.581 $\pm$ 0.012 \\
        \midrule
        \multirow{3}{*}{\shortstack[l]{Window Percentage\\($\gamma$)}} 
        & 0.5 & 79.21 $\pm$ 0.41\% & 0.559 $\pm$ 0.010 \\
        & 0.8 (Base) & 78.65 $\pm$ 0.72\% & 0.585 $\pm$ 0.019 \\
        & 1.0 & 77.68 $\pm$ 0.72\% & 0.594 $\pm$ 0.022 \\
        \midrule
        \multirow{3}{*}{\shortstack[l]{Recovery Steps\\($R$)}} 
        & 1 & 77.71 $\pm$ 0.81\% & 0.602 $\pm$ 0.025 \\
        & 3 (Base) & 78.65 $\pm$ 0.72\% & 0.585 $\pm$ 0.019 \\
        & 5 & 79.78 $\pm$ 0.35\% & 0.551 $\pm$ 0.009 \\
        \midrule
        \multirow{3}{*}{\shortstack[l]{Latent Dimension}} 
        & 8 & 79.17 $\pm$ 0.55\% & 0.574 $\pm$ 0.007 \\
        & 16 (Base) & 78.65 $\pm$ 0.72\% & 0.585 $\pm$ 0.019 \\
        & 32 & 78.18 $\pm$ 0.73\% & 0.599 $\pm$ 0.024 \\
        \midrule
        \multirow{3}{*}{\shortstack[l]{Hidden Dimension\\($h$)}} 
        & 64 & 79.14 $\pm$ 0.51\% & 0.576 $\pm$ 0.021 \\
        & 128 (Base) & 78.65 $\pm$ 0.72\% & 0.585 $\pm$ 0.019 \\
        & 256 & 79.10 $\pm$ 0.37\% & 0.580 $\pm$ 0.015 \\
        \midrule
        \multirow{3}{*}{\shortstack[l]{AE Epochs\\($e$)}} 
        & 5 & 78.85 $\pm$ 0.29\% & 0.582 $\pm$ 0.019 \\
        & 10 (Base) & 78.65 $\pm$ 0.72\% & 0.585 $\pm$ 0.019 \\
        & 20 & 78.98 $\pm$ 0.91\% & 0.584 $\pm$ 0.010 \\
        \midrule
        \multirow{2}{*}{\shortstack[l]{AE Architecture}} 

        & Residual (Base) & 78.65 $\pm$ 0.72\% & 0.585 $\pm$ 0.019 \\
        & Progressive & 78.56 $\pm$ 1.11\% & 0.586 $\pm$ 0.028 \\
        \bottomrule
    \end{tabular}
    }
    \caption{Sensitivity analysis on the FMNIST dataset (Dirichlet $\alpha=0.5$) over three random seeds, varying one hyperparameter at a time from the base configuration. Mean $\pm$ standard deviation across seeds of the test accuracy and generalisation loss, each averaged over the final $10\%$ of rounds (Table~\ref{tab:rounds}).}
    \label{tab:sensitivity_analysis}
\end{table*}

\section{Hyperparameter Sensitivity Analysis}
\label{app:sensitivity_analysis}
 
To assess \textsc{FedOAED}'s sensitivity to hyperparameter selection, we report the generalisation performance across a one-dimensional grid search. For these experiments, we varied one parameter at a time while keeping the others fixed at their base values (utilising the Residual autoencoder architecture for the sweep).
 
The results, detailed in Table~\ref{tab:sensitivity_analysis}, indicate that \textsc{FedOAED} is stable across the range examined. Every configuration attains between $77.68\%$ and $79.78\%$, against a FedAvg baseline of $75.28 \pm 0.81\%$ on the same partition (Table~\ref{tab:ablation_safety}). The weakest setting we tested still exceeds the baseline by more than two points, indicating that the safety mechanisms prevent performance degradation even under sub-optimal hyperparameters. At three seeds the recovery-step, window-percentage and mixing-coefficient columns separate from seed variance; the remaining columns span differences comparable to it and should be treated as unresolved.

\subsection{Key Observations}

The sweep highlights three dynamics of the on-device denoising process. First, the recovery-step budget produces the widest span: $R = 5$ yields the best result in the table ($79.78 \pm 0.35\%$, loss $0.551 \pm 0.009$) and $R = 1$ among the worst ($77.71 \pm 0.81\%$), a span of $2.07$ points against standard deviations below $0.9$. Second, the mixing coefficient responds monotonically over the range tested, in both accuracy ($77.84 \to 79.45\%$) and loss ($0.602 \to 0.567$), saturating between $0.05$ and $0.10$; we discuss the significance of this below. Third, window length runs in the direction of shorter trajectories: $\gamma = 0.5$ attains $79.21 \pm 0.41\%$ against $77.68 \pm 0.72\%$ at $\gamma = 1.0$, consistent with the concern that long windows draw the reconstruction towards stale, high-loss states. Autoencoder capacity, by contrast, is not a limiting factor: the two architectures are indistinguishable ($78.65$ against $78.56\%$), the smallest latent width outperforms the largest, and halving the hidden width costs nothing. Although these differences fall within three-seed variance, larger autoencoder capacities show no clear benefit.

\subsection{Reading the Mixing-Coefficient Column}\label{app:sweep_reading}

The mixing coefficient is the parameter that most directly controls how much of the reconstruction reaches the model: by Eq.~\eqref{eq:convex_combination} the parameter change is proportional to $c\lambda$, so $\lambda$ scales the refinement whilst leaving the rest of the local procedure (the snapshot buffer, the autoencoder, the gates and the $R$ recovery steps) unchanged. Its column is therefore the closest the sweep comes to isolating the contribution of the denoising itself.

Both metrics respond monotonically across the range tested: accuracy rises from $77.84 \pm 0.90\%$ at $\lambda = 0.001$ to $79.45 \pm 0.51\%$ at $\lambda = 0.05$, and loss falls from $0.602$ to $0.567$ over the same interval, with a plateau between $0.05$ and $0.10$ that is well within seed variance. A dose-response relationship of this kind is difficult to attribute to the additional local optimisation the refinement cycle introduces, since the recovery-step budget is held fixed at $R = 3$ throughout the column, though the realised number of steps varies with the gate's acceptance rate.

Because the sweep omits $\lambda = 0$, these trends should be interpreted as supporting evidence rather than a strict ablation; separating the refinement effect from the extra recovery steps would require a step-matched FedAvg baseline with $\tilde{K} = K + JR$ steps (Appendix~\ref{app:ablation_caveats}). The sweep is also one-dimensional and anchored at a single configuration, so it establishes local rather than global robustness and cannot detect interactions between hyperparameters.

\section{Hyperparameter Configuration and Tuning Guidelines}
\label{app:tuning_guidelines}
 
Given the non-standard inclusion of an autoencoder directly within the client optimisation loop, we provide explicit guidelines for configuring \textsc{FedOAED} in practice. The recommendations follow from the measurements of Appendices~\ref{app:compute_profile} and~\ref{app:sensitivity_analysis}.
 
\subsection{Autoencoder Architecture and Scaling}
\begin{itemize}
    \item \textbf{Chunk Percentage ($\phi$):} Controls the trade-off between denoiser size and global context. Smaller chunks (e.g., $0.01$) reduce the autoencoder's parameter count proportionately, but restrict the model from learning correlations between distant neural network layers. We use $\phi = 0.01$ throughout and did not sweep it; values in $[0.01, 0.05]$ are a reasonable starting range for backbones of the scale considered here. Note two properties: autoencoder training FLOPs are invariant in $\phi$ (Appendix~\ref{app:compute_profile}), and because $C = \phi d$ the autoencoder still grows linearly in $d$, so $\phi$ must be reduced as the backbone grows, or the chunk width fixed in absolute terms, if the denoiser is to stay small (Appendix~\ref{app:ae_scaling}).
 
    \item \textbf{Latent Dimension:} Determines the compression bottleneck. Because the autoencoder operates on a short, single-client trajectory, a bottleneck is preferable. In our sweep, accuracy decreased monotonically from latent dimension $8$ to $32$, suggesting that wider bottlenecks overfit to stochastic gradient noise rather than filtering it.
 
    \item \textbf{AE Epochs ($e$):} Applies only to the first refinement cycle of a round; every subsequent cycle fine-tunes the persistent autoencoder for a single epoch. Total AE training per round is therefore $e + (\lfloor E/E_r \rfloor - 1)$ epochs rather than $e$ per cycle, and values beyond $\approx 5$ have limited effect, consistent with Table~\ref{tab:sensitivity_analysis}.
\end{itemize}
 
\subsection{Trajectory Buffer and Loss Weighting}
\begin{itemize}
    \item \textbf{Snapshot Window ($\gamma$):} Dictates how much history the autoencoder processes. If too large, the autoencoder reconstructs stale, high-loss parameter states from early in the epoch. If too small, it lacks the data volume to learn a meaningful manifold. Setting $\gamma$ to encompass roughly $50\%$ to $80\%$ of the epoch batches strikes a reasonable balance; in our sweep $\gamma = 0.5$ was the strongest setting and $\gamma = 1.0$ the weakest, so shorter windows are preferable both for accuracy and where memory is constrained. Ensure $W \ge m$: on clients with very few batches per epoch an unclamped window leaves the element-wise deviation undefined (Appendix~\ref{app:divergences}).
 
    \item \textbf{Snapshot Interval ($s$) and Buffer Capacity ($Q$):} The two direct controls on buffer memory, which scales as $\mathcal{O}(\min(Q, E b_i / s)\, d)$. Raising $s$ subsamples the trajectory at proportionate saving and is the first parameter to adjust when moving to a larger backbone; $Q$ caps the term outright and is the binding control on datasets with many local batches, as on Shakespeare (Appendix~\ref{app:memory_footprint}).
 
    \item \textbf{Loss Temperature ($\kappa$):} Controls the sharpness of the softmax weighting over trajectory snapshots. Setting $\kappa \to 0$ collapses training onto the single best snapshot, turning the autoencoder into an identity mapping of the minimum. Setting $\kappa \to \infty$ treats all snapshots equally, ignoring the loss landscape entirely. A base value of $1.0$ effectively bounds the weight skew.
\end{itemize}
 
\subsection{Refinement and Safety Gates}
\begin{itemize}
    \item \textbf{Mixing Coefficient ($\lambda$):} The most critical hyperparameter, bounding the maximum allowed perturbation size. Accuracy increases monotonically in $\lambda$ up to $0.05$ and plateaus thereafter, even under Dirichlet skew (Appendix~\ref{app:sweep_reading}), so $\lambda \approx 0.05$ is the empirically preferred setting. We report $\lambda = 0.01$ as a conservative default, trading roughly $0.8$ accuracy points for a smaller worst-case parameter change; practitioners prioritising accuracy should prefer $0.05$, and values below $0.005$ are not recommended, as the refinement becomes too small to have an effect.
 
    \item \textbf{Relative Threshold ($\tau$):} The absolute cutoff for the relative reconstruction error. A threshold of $0.3$ safely filters out untrained or collapsed autoencoders while allowing meaningful structural reconstructions to pass.
 
    \item \textbf{Loss Tolerance ($\rho$):} The rollback gate's strictness. Although the gate is deterministic on $\widehat{\mathcal{D}}_i$, those $b$ batches are a small subsample of the local data, so an improvement measured on them need not track the full local objective. Setting $\rho = 0.0$ is excessively rigid and rejects most refinements. A tolerance of $0.05$ to $0.1$ allows the refinement to step slightly out of a sharp local minimum to find a flatter basin, aiding generalisation.
 
    \item \textbf{Recovery Steps ($R$):} The widest-spanning parameter in our sweep, with $R = 5$ giving the strongest result overall. Recovery steps are nonetheless ordinary SGD steps, and at $E_r = 1$ they already add fifty per cent to the local budget on FMNIST, so increases should be reported alongside the equivalent FedAvg step count to keep the two effects distinct (Appendix~\ref{app:ablation_caveats}).
\end{itemize}

\section{Limitations}
\label{app:limitations}

Section~\ref{sec:conclusion} states the principal trade-off compactly. We set out the full picture here, grouped into constraints on deployment and limits of the empirical evidence. The scope of the convergence guarantee is stated in Section~\ref{sec:theory}.
 
\subsection{Deployment Constraints}

\begin{itemize}
    \item \textbf{On-device computation.} A local update takes approximately three times the wall-clock time of FedAvg (Appendix~\ref{app:compute_profile}). We argue this is favourable in cross-device federated learning, where communication latency and server scalability bind before client FLOPs and where the cost falls on the sampled subset rather than the population. The argument does not transfer to settings in which clients are compute-bound or energy-constrained, nor to cross-silo deployments where participation is near-complete and the per-round cost is therefore incurred by every client at every round.
 
    \item \textbf{Autoencoder size on large backbones.} Because the chunk width is specified as a fraction of the model dimension, $C = \lfloor \phi d \rfloor$, the denoiser's parameter count remains linear in $d$ despite chunking. Under the configuration reported here the autoencoder is comparable to or larger than the backbone it denoises (Appendix~\ref{app:ae_scaling}). Although chunking reduces the projection layers by a factor of $\phi$ and expands the training set by $1/\phi$, scaling to substantially larger models would require decreasing $\phi$ or fixing an absolute chunk width. The effect is large: fixing $C = 1024$ irrespective of $d$ would reduce the ResNet-18 denoiser from $\approx\!5.8\times10^{7}$ parameters to $\approx\!1.6\times10^{6}$, a thirty-sevenfold reduction, and would make the count independent of the backbone entirely. We have not evaluated the accuracy consequences of either choice, and regard this as the most direct route to scaling the method.
 
    \item \textbf{Trajectory buffer.} The buffer holds up to $Q$ flattened parameter vectors and is likewise linear in $d$. It is immaterial for the two $6\times10^{4}$-parameter models we study, but reaches the order of gigabytes for the ResNet-18 configuration (Appendix~\ref{app:memory_footprint}). Our CIFAR-10 results were obtained in simulation, where this is unproblematic; on physical mobile hardware it would require raising the snapshot interval $s$, lowering $Q$, or storing snapshots at reduced precision. We have not measured the accuracy cost of buffer compression, and doing so is the most direct route to deployment on constrained hardware.
 
    \item \textbf{Statelessness.} Discarding $\mathcal{A}_i$, the buffer, and the window statistics after each round avoids stale client state under $1$--$2\%$ participation. The cost is that no denoising prior accumulates across rounds, and each visit must relearn one from a short trajectory. A mechanism for sharing priors across clients without introducing per-client server state would relax this, and is outside the scope of the present paper.
 
    \item \textbf{Interaction with privacy mechanisms.} \textsc{FedOAED} operates on local parameters and transmits only a pseudo-gradient, so it is in principle compatible with secure aggregation and with differentially private client updates. We have not analysed that composition. In particular, the interaction between the refinement and the sensitivity bound required for a differential privacy guarantee is unexamined, and we make no privacy claim beyond that of standard FedAvg.
\end{itemize}

\subsection{Limits of the Empirical Evidence}

\begin{itemize}
    \item \textbf{Benchmark and architecture coverage.} We evaluate FMNIST with LeNet-5, CIFAR-10 with ResNet-18, and Shakespeare with a GRU. This spans two modalities and three architectures, but omits the larger vision and language backbones now common in the field, and both image benchmarks use synthetic partitions; only Shakespeare is naturally partitioned. Whether the trajectory of a substantially larger model retains enough low-dimensional structure for a chunk-level denoiser to exploit is an open empirical question, and one made harder to answer by the memory scaling discussed above.
 
    \item \textbf{Seeds and reported variance.} Standard deviations are reported for FMNIST but not for CIFAR-10 or Shakespeare, and the sensitivity sweep is single-anchored. Differences of under one percentage point on those two datasets should not be interpreted as significant, and the CIFAR-10 LQ-3 result in particular, where FedProx leads \textsc{FedOAED} by $0.80$ points, falls within a range that repeated runs might reorder.
 
    \item \textbf{Ablation and sweep design.} The ablations do not isolate the recovery steps from the refinement itself (Appendix~\ref{app:ablation_caveats}). The sweep is one-dimensional, anchored at a single configuration, and cannot detect interactions between hyperparameters.
 
    \item \textbf{Baseline reproduction.} MIFA and FedVARP were reproduced independently, in the variant using a fixed client learning rate rather than a server optimiser. MIFA in particular is unstable under our participation rates, and whilst this is consistent with the mechanism we identify, stale cached updates injected on behalf of absent clients, we cannot exclude that a more favourable configuration would narrow the gap. FedVARP outperforms \textsc{FedOAED} on Shakespeare at both learning rates, by $0.07$ and $0.51$ points, and FedProx does so on CIFAR-10 LQ-3 by $0.80$; all three are reported. Shakespeare is single-seed, so these differences are not resolved.
 
    \item \textbf{Hyperparameter tuning.} We use standard benchmark hyperparameters for CIFAR-10 and Shakespeare rather than tuning per method. While this ensures a uniform setup across baselines, per-method tuning could alter the remaining gaps on those datasets.
\end{itemize}


\end{document}